\documentclass{article}

\usepackage{arxiv}

\usepackage{url}            
\usepackage{amsfonts}       
\usepackage{nicefrac}       
\usepackage{microtype}      
\usepackage{fancyhdr}       

\usepackage{multirow}
\usepackage{hyperref}
\usepackage{float}
\usepackage{verbatim} 
\restylefloat{figure}
\usepackage{caption}
\usepackage{subcaption}
\usepackage[font=small,skip=0pt]{caption}
\usepackage{booktabs}
\usepackage{longtable}
\usepackage{tabularx}
\usepackage{cancel}
\usepackage{multirow}
\usepackage{supertabular}
\usepackage{algorithmic}
\usepackage{algorithm}
\usepackage{amsmath}
\usepackage[normalem]{ulem}
\usepackage[utf8]{inputenc}
\usepackage[T1]{fontenc}
\usepackage{lineno}
\usepackage{array}
\usepackage{makecell}
\usepackage{pdflscape}
\usepackage{afterpage}
\usepackage{capt-of}
\usepackage{lipsum}
\usepackage{changepage}
\usepackage{xcolor}
\usepackage{rotating} 
\usepackage{amssymb}
\usepackage{multicol}

\usepackage{graphicx}
\usepackage{svg}

\usepackage{colortbl}
\usepackage{tabularray}
\usepackage{braket}
\usepackage[export]{adjustbox}

\usepackage[shortlabels]{enumitem}

\title{Bidirectional recurrent imputation and abundance estimation of LULC classes with MODIS multispectral time series and geo-topographic and climatic data}

\author{
  José Rodríguez-Ortega \\
  Dept. of Computer Science and Artificial Intelligence, DaSCI \\
  University of Granada \\
  Granada \\
  \texttt{e.jrodriguez98@go.ugr.es} \\
  LifeWatch-ERIC ICT Core \\
  Seville \\
  \texttt{jose.rodriguez@lifewatch.eu} \\
  \And
  Rohaifa Khaldi \\
  LifeWatch-ERIC ICT Core \\
  Seville \\
  \texttt{rohaifa@go.ugr.es} \\
   \And
  Domingo Alcaraz-Segura \\
  Dept. of Botany \\
  Faculty of Science, University of Granada \\
  Granada\\
  \texttt{dalcaraz@ugr.es} \\
  \And
  Siham Tabik \\
  Dept. of Computer Science and Artificial Intelligence, DaSCI \\
  University of Granada \\
  Granada\\
  \texttt{siham@ugr.es} \\
}

\begin{document}
\maketitle

\begin{abstract}
Remotely sensed data are dominated by mixed Land Use and Land Cover (LULC) types. Spectral unmixing (SU) is a key technique that disentangles mixed pixels into constituent LULC types and their abundance fractions. While existing studies on Deep Learning (DL) for SU typically focus on single time-step hyperspectral (HS) or multispectral (MS) data, our work pioneers SU using MODIS MS time series, addressing missing data with end-to-end DL models. Our approach enhances a Long-Short Term Memory (LSTM)-based model by incorporating geographic, topographic (geo-topographic), and climatic ancillary information. Notably, our method eliminates the need for explicit endmember extraction, instead learning the input-output relationship between mixed spectra and LULC abundances through supervised learning. Experimental results demonstrate that integrating spectral-temporal input data with geo-topographic and climatic information significantly improves the estimation of LULC abundances in mixed pixels. To facilitate this study, we curated a novel labeled dataset for Andalusia (Spain) with monthly MODIS multispectral time series at 460m resolution for 2013. Named Andalusia MultiSpectral MultiTemporal Unmixing (Andalusia-MSMTU), this dataset provides pixel-level annotations of LULC abundances along with ancillary information. The dataset \footnote{\url{https://zenodo.org/records/7752348}} and code \footnote{\url{https://github.com/jrodriguezortega/MSMTU}} are available to the public.
\end{abstract}


\section{Introduction} \label{sec:intro}
\begin{figure*}[h!]
\centering
\begin{subfigure}{0.46\textwidth}
    \includegraphics[width=\textwidth]{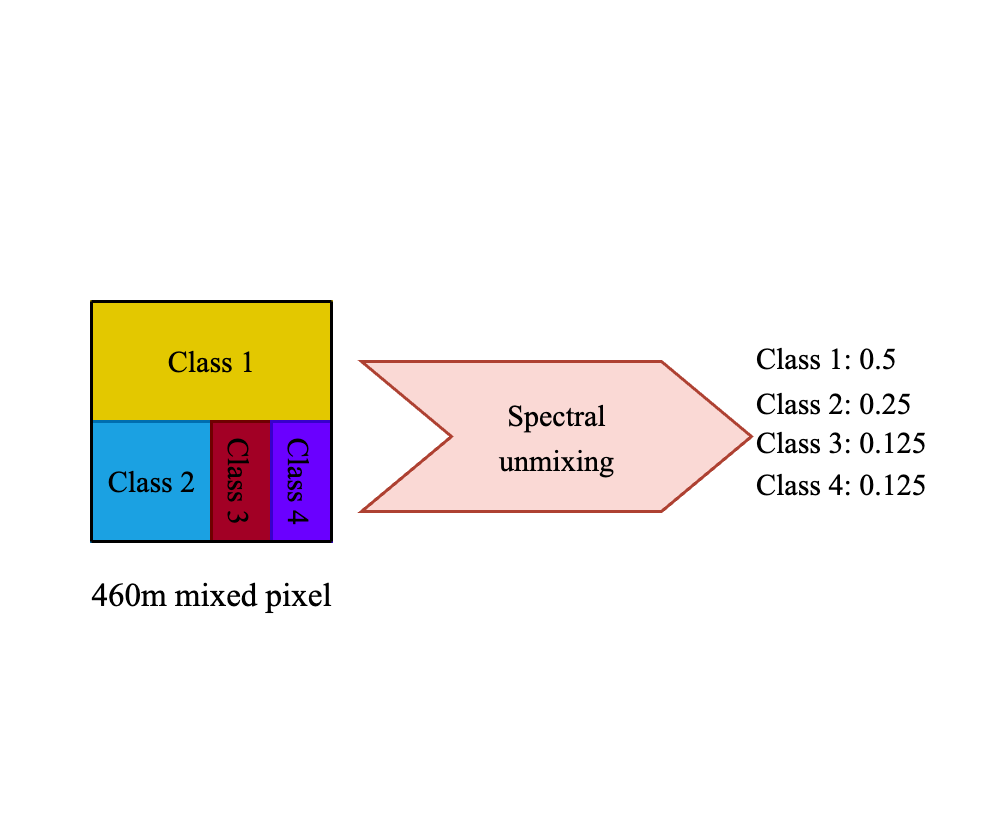}
    \caption{Spectral unmixing without endmember extraction.}
    \label{fig:first}
\end{subfigure}
\hspace{1cm}
\begin{subfigure}{0.46\textwidth}
    \includegraphics[width=\textwidth]{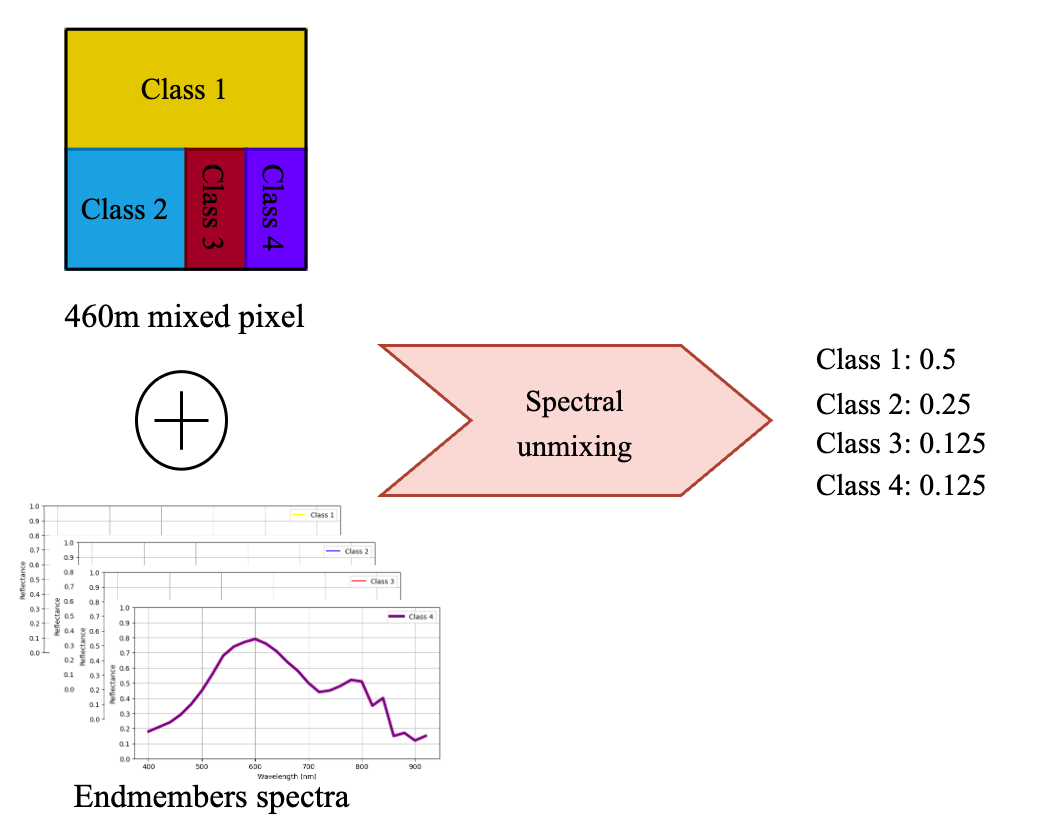}
    \caption{Spectral unmixing with endmember extraction.}
    \label{fig:second}
\end{subfigure}

\vspace*{+2mm}
\caption{Toy example illustrating (a) Spectral unmixing without endmember extraction versus (b) Spectral unmixing with endmember extraction in a $460m$ mixed pixel.}
\label{fig:su_toy}
\end{figure*}

LULC mapping is normally addressed by classifying each pixel in a satellite image into a LULC class, also known as semantic segmentation (SS) in RS images. Frequently, the spatial resolution of an image and the thematic resolution of its LULC legend do not match, which leads to the mixed pixel problem, where pixels are not pure but contain several LULC classes. Accordingly, many methods have tried to estimate the relative abundances of each LULC class in a pixel from the combined spectral signature \cite{schowengerdt2006remote}. Such estimation of the spectrum and the abundance of the LULC classes present within each pixel is known as Spectral Unmixing (SU) and is one of the most challenging areas of research in Remote Sensing (RS) \cite{zhang2018hyperspectral}. Various unmixing approaches, including linear and nonlinear methods, have been developed \cite{keshava2002spectral, 9324546}. Many of these approaches require the use of the pure spectral signature (the endmember) of each LULC class. However, the acquisition of endmembers might be hard in areas dominated by mixed pixels \cite{wang2021spatio}.
To overcome this limitation, several methods have been introduced to avoid the need of endmembers extraction \cite{licciardi2011pixel, zhang2018hyperspectral, wang2021spatio, wang2021real} as depicted in Figure \ref{fig:su_toy}.

In the last years, modern DL models have been increasingly employed for addressing SU by directly learning the input-output mapping from the spectra of mixed pixels to their corresponding class abundances. Several studies explored the potential of DL methods for SU in LULC mapping using either single time-step HS data \cite{zhang2018hyperspectral, palsson2018hyperspectral, ghosh2022hyperspectral} or single time-step MS data \cite{mustard1998nonlinear}. Including temporal information could be a great opportunity to improve SU methods \cite{9324546} and a few works (see Table \ref{table:2}) have started exploring approaches with MS time series data. However, to the best of our knowledge, none have explored an end-to-end DL solution, where recurrent neural networks (RNNs) and LSTM networks are a perfect fit. 

In contrast to traditional methods, the application of DL in SU facilitates the exploitation of ancillary information such as geographic location, topography, and climate. For example, in the field of computer vision, ancillary data has been successfully used by DL models to improve the performance during image classification \cite{berg2014birdsnap, ellen2019improving, tseng2022timl}. However, the introduction of ancillary information remains unexplored in spectral unmixing methods.  We hypothesize that injecting such ancillary information could boost the performance of the predictive model in spectral unmixing. This information may help the model understand the spatial distribution and variations in climate of the different LULC types.

The primary problem addressed in this study is the spectral unmixing of LULC classes using MS time series data and ancillary information, and it faces several challenges:
\begin{itemize}
    \item Public labeled datasets with MS multitemporal data for spectral unmixing of LULC classes are not available. 
    
    \item Creating a new dataset of MS time series  plus ancillary information together with LULC abundances annotations is complex, costly and time consuming.
    
    \item Remote sensing data usually contains missing values due to atmospheric conditions or sensors' errors, which requires applying robust processing techniques.

    \item Feeding ancillary information to spectral unmixing models is a promising direction but can be complex. Ensuring that these data improve the model robustness is a challenge and it is not explored yet.
\end{itemize}

\begin{figure*}[h!]
    \centering
    \includegraphics[width=0.8\textwidth]{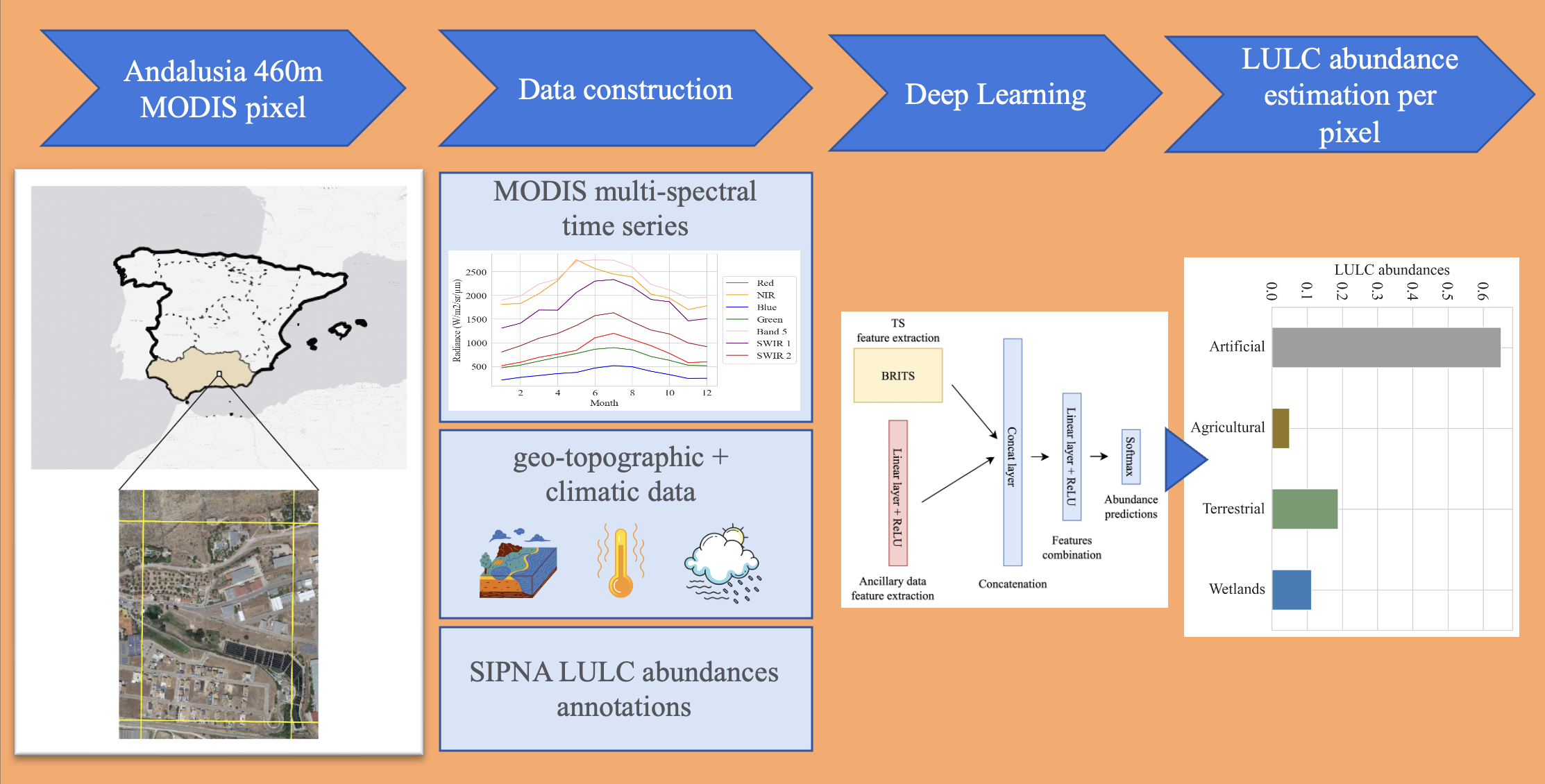}
    \caption{Flowchart of our proposed method. First, Andalusia-MSMTU dataset is built using MODIS MS multitemporal data plus geo-topographic and climatic data together with the corresponding LULC abundances annotations extracted from SIPNA. Subsequently, the deep learning based model is designed to use both, multi-spectral times series and geo-topographic and climatic data to estimate the LULC abundances. }\
    \label{fig:grabs}
\end{figure*}

Given the above mentioned challenges, the main objective of this study is twofold: (1) to create a regional-scale dataset of more than $500,000$ MODIS $460m$ resolution pixels from Andalusia, Spain, and (2) to develop DL-based approach for SU, without the need of endmembers extraction, that estimates the LULC abundances per pixel using MS time series and ancillary data. This dataset provides for each individual pixel: (a) a MS time series of monthly observations during the year 2013 of the seven spectral bands of the MODIS sensor, (b) ancillary information containing geographic, topographic and climatic variables, and (c) their corresponding LULC class abundances at two different levels of the classes hierarchy, extracted Andalusia's official LULC map (SIPNA \cite{SIPNA}). Furthermore, the DL-based method consists of a two branch neural network (NN) where the first branch process the MS time series using a LSTM-based model capable of handling missing values, and the second branch process the ancillary information.  A graphical illustration of the followed workflow in this study is shown in Figure \ref{fig:grabs}.

Two assumptions are made in this study: (1) LULC changes within a one-year timeframe are limited at a 460m pixel resolution, so our LULC abundance annotations are assumed to be static. (2) The selected MODIS time series data, ancillary information, and LULC annotations adequately represent the land dynamics of Andalusia, since more than $500,000$ pixels from Andalusia are collected, representing almost the whole region. The constraint includes the challenge of dealing with missing values in remote sensing data, which we solve by proposing a DL method capable of handling missing values. 

The motivation behind this research is rooted in the need for improved methods to perform SU in complex, heterogeneous landscapes with MS time series data. The absence of accessible labeled datasets, combined with the complexity of creating new datasets, underscores the significance of developing innovative approaches to advance the field.

The primary contributions of this research can be summarized as follows: 
\begin{itemize}
    \item We built Andalusia-MSMTU dataset: a novel MS multitemporal labeled dataset with mixed pixels from Andalusia, a highly heterogeneous region in Spain. Each pixel is annotated with LULC abundances. In addition to the MS multitemporal information, each mixed pixel has its corresponding geo-topographic and climatic   information. Such dataset will open the possibility for new explorations.
    
    \item We designed and analyzed a DL-based approach that estimates the LULC abundances per pixel of LULC classes from MS time series data with and without ancillary information.
    
\end{itemize}

The rest of this article is structured as follows: Section \ref{sec:related} presents the related work. Section \ref{sec:back} provides preliminaries and background. Section \ref{sec:study} introduces the study area and the data construction process. Section \ref{sec:metho} describes the used DL methodology. Section \ref{sec:results} assesses the results obtained. Section \ref{sec:discussion} provides a comprehensive discussion of the obtained results, comparing them with previous works. Finally, section \ref{sec:conclusions} concludes the findings and sheds light into future works.

\section{Related work}\label{sec:related}
Firstly, general DL in RS methods are reviewed. Subsequently, related works on spectral unmixing overall, with a specific focus on employing deep learning methodologies, are introduced. Finally, works that build labeled datasets designed for the unmixing approaches are reviewed and comprehensively summarized in Table \ref{table:2}.

\subsection{DL in RS}
Thanks to the recent success of DL methods in many learning tasks, tons of efforts have been made to bring DL to RS field \cite{yuan2020deep}. Concretely, LULC classification task is of paramount importance since many environmental applications rely on LULC maps, such as urban planning, forest monitoring, change detection...

Traditionally, only one source of input data was used to perform the classification task, that is only using HS \cite{8697135}, MS \cite{kemker2018algorithms}, LiDAR \cite{guan2015deep} or synthetic aperture radar (SAR) \cite{lv2015urban}. Recently, multimodal models has emerged with the promise to improve the LULC classification by combinaning the different input data types. \cite{9174822} addresses challenges in LULC classification using a multimodal deep learning (MDL) framework. It tackles limitations of traditional deep learning in complex scenes, introducing five fusion architectures and emphasizing applicability beyond pixel-wise classification to spatial information modeling. Also, \cite{9724217} introduced MUNet, a multimodal unmixing network for HS images, leveraging LiDAR data to enhance discrimination in complex scenes. MUNet uses a SE-driven attention mechanism, incorporating height differences from LiDAR for improved performance. \cite{8954957} presented IISU, an illumination invariant spectral unmixing model addressing spectral variability caused by variable incident illuminations. Utilizing radiance HS data and a LiDAR-derived digital surface model, IISU provides explicit explanations for endmember variability, outperforming existing models, particularly in shaded pixels. The proposed model yields more accurate abundances and shadow-compensated reflectance. In \cite{hong2023cross}, authors built the C2Seg dataset for cross-city LULC classification, addressing limitations of DL models across diverse urban environments. Their proposed HighDAN network, employing high-resolution domain adaptation and adversarial learning, demonstrates superior segmentation performance and generalization abilities compared to existing methods. Following the modern self-supervised learning (SSL) paradigm, SpectralGPT \cite{hong2023spectralgpt} is proposed as a novel universal foundation model tailored for spectral remote sensing data, utilizing a 3D generative pretrained transformer. Trained on one million spectral RS images, it accommodates varied inputs, leverages 3D token generation for spatial-spectral coupling, and achieves substantial performance gains across geoscience tasks like scene classification and semantic segmentation. Finally, \cite{hong2023decoupled} introduces a subpixel-level HS super-resolution framework, DC-Net, addressing the distribution gap between HS and high spatial resolution MS images. The novel decoupled-and-coupled network progressively fuses information from pixel to subpixel-level, mitigating spatial and spectral resolution differences. Employing a SSL module ensures material consistency for enhanced HS restoration.

\subsection{Spectral unmixing}

The existing spectral unmixing methods can be broadly categorized as linear mixture models (LMM) and nonlinear mixture models (NLMM) according to the formulation describing the underlying mixing process of endmembers \cite{heylen2014review}. 

LMM consider that the spectral signature of a mixed pixel is a weighted sum of the endmember spectra and that the weights associated with the endmembers are given by their corresponding relative area abundance in the pixel. LMM-based methods have been widely developed in last decades including linear, geometrical, nonnegative matrix factorization, bayesian and fuzzy models among others \cite{heinz2001fully, keshava2002spectral, quintano2012spectral, bioucas2012hyperspectral, ghaffari2017reducing, 8747263}. LMM typically assumes that the spectrum of each LULC class is characterized by a single fixed endmember. However, pure pixels from the same LULC class may have different spectra, which is called intra-class variability \cite{zhang2019assessing}. To overcome this limitation, several multiple endmember spectral mixture analysis (MESMA) models have been developed \cite{roberts1998mapping, bateson2000endmember, li2016geostatistical, degerickx2019enhancing}.

Since the extraction of a large number of pure endmembers is still a great challenge in areas dominated by mixed pixels, several works without assuming any prior knowledge about the mixing process were introduced. These methods, also known as blind spectral unmixing (BSU) methods, include independent component analysis \cite{wang2006applications, moussaoui2008decomposition, xia2011independent}, non-negative matrix factorization \cite{miao2007endmember, yang2010blind, zhu2014structured, feng2022hyperspectral},  sparse component analysis \cite{zhong2016blind} or wavelet-based \cite{9555396} methods.

Given the nonlinear mixing effects of endmembers, NLMM have been proposed through the years to overcome LMM limitations and enhance the spectral unmixing performance. These include bilinear models \cite{halimi2011nonlinear}, radial basis function networks \cite{altmann2011nonlinear}, kernel-based models \cite{broadwater2007kernel}, neural networks and low-rank tensor \cite{9386217} methods among others.

\subsubsection{DL in spectral unmixing}

Spectral unmixing have also met DL models, which fall under the category of NLMM. One of the first DL approaches for SU was proposed by \cite{foody1996relating}, where they introduced three spectral bands values and the neural network (NN) predicts the abundances of three LULC classes. \cite{atkinson1997mapping} compared NN, Linear Mixture Models (LMM), and fuzzy c-means for spectral unmixing of LULC classes, being the NN the best model given sufficient training samples. Then, \cite{licciardi2011pixel} proposed a two-stage NN architecture that first reduces the dimension of the input vector using an auto-associative NN, and performs abundance estimation out of the reduced input using a MLP. Recently, \cite{palsson2018hyperspectral} evaluated autoencoders with different hyperparameters. \cite{yu2022multi} introduced MSNet, a multi-stage convolutional autoencoder network designed for linear HU, achieving this by capturing contextual relationships between pixels. \cite{9383423} introduced CyCU-Net for HU, enhancing performance by incorporating cycle consistency and self-perception loss. The network, leveraging cascaded autoencoders, preserves detailed material information and achieves high-level semantic preservation during unmixing. \cite{9399660} introduced SeCoDe, a novel blind HS unmixing model designed for airborne and spaceborne HS imagery. Leveraging sparsity-enhanced convolutional decomposition, SeCoDe effectively addresses spectral variabilities and maintains continuous spectral components. Going beyond autoencoder-like architectures, \cite{9160879} introduced Deep HSNet, a novel siamese network for HU that considers diverse endmember properties from different extraction algorithms. Deep HSNet incorporates a subnetwork to effectively learn endmember information, enhancing the accuracy of the unmixing process. Following the success of transformers architecture \cite{vaswani2017attention}, \cite{ghosh2022hyperspectral} and \cite{li2023attention} introduced NN architectures with the attention mechanism for abundance estimation. Regarding SSL for spectral unmixing works, \cite{9610077} proposed a two-stage fully connected SSL network for BSU, addressing challenges of limited supervision and data requirements. The network jointly estimates endmembers and abundances in the first stage, and learns HS image acquisition physics in the second stage. Also, AutoNAS \cite{9807268} explored neural architecture search (NAS) for determining optimal network architecture in HU. Utilizing SSL and an affine parameter sharing strategy, it achieves optimal channel configuration. Further, an evolutionary algorithm enables flexible convolution kernel search.

Regarding RNN-based works, the only work on SU using a LSTM-based network was introduced by \cite{zhao2021lstm}. They proposed a nonsymmetric autoencoder network with a LSTM component to capture spectral correlation together with an attention mechanism to further enhance the unmixing performance. For a more detailed review of DL methods in spectral unmixing see \cite{heylen2014review} and \cite{9324546}.

In parallel, there exist few works that incorporate ancillary data to improve the performance of DL models. Most of these studies occur in the field of computer vision, such as high inter-class similarity classification problems \cite{berg2014birdsnap}, plankton image classification \cite{ellen2019improving} or crop type mapping \cite{tseng2022timl}.

\renewcommand{\arraystretch}{1.5}
\begin{table*}[h!]
\centering
\caption{List of spectral unmixing works using MS and multitemporal data. HRM: Higher Resolution Map, PTU: Probabilistic Temporal Unmixing, FCLSU: Fully Constrained Least Squares Spectral Unmixing, TMA: Temporal Mixture Analysis, RF: Random Forest, OLS: Ordinary Least Square, STSU: Spatio-Temporal Spectral Unmixing, RSTSU: Real-time STSU, LSTM: Long-Short Term Memory.}
\vspace*{+1mm}
\resizebox{\textwidth}{!}{\begin{tabular}{l l l r l r l l} 
 
 Ref. \& year & Study area  & Source & Spatial-temporal res. & Annotations & \# classes & Public & Approach \\ \hline 
 \multirow{2}{4em}{\cite{lobell2004cropland} 2004} & Yaqui Valley  (Mexico) & MODIS & 500m - 8 obs. (months) & HRM (Landsat-derived) & 3 & No & PTU \\ 
  & Great Plains (US) & MODIS & 500m - 8 obs. (months) & HRM (Landsat-derived) & 3 & No & PTU \\ \hline
 \multirow{2}{4em}{\cite{zurita2011multitemporal} 2011} &  The Netherlands & MERIS & 300m - 7 obs. (months) & HRM (LGN5 L1) & 4 & No & FCLSU  \\
 
 & The Netherlands & MERIS & 300m - 7 obs. (months) & HRM (LGN5 L2) & 12 & No & FCLSU  \\ \hline
 \cite{yang2012temporal} 2012 & Japan & MODIS & 250m - 365 obs. (days) & HRM (Landsat-derived) & 2 & No & TMA-based  \\ \hline
 \cite{deng2020continuous} 2020 & Broome County (US)& Landsat  & 30m - (2000 to 2014) & HRM (NLCD) & 2 & No & RF \\ \hline
 \cite{bullock2020monitoring} 2020 & Rondônia (Brazil) & Landsat  & 30m - (1990 to 2013) & HRM (Rondônia) & 5 & No & OLS + RF \\  \hline 

\multirow{3}{4em}{\cite{wang2021spatio} 2021} & Daxing (China)& MODIS & 480m - 3 obs. (months) & HRM (Landsat-derived) & 2 & No & STSU \\
 & Zibo (China)& MODIS & 480m - 3 obs. (months) & HRM (Landsat-derived) & 2 & No & STSU\\ 
 & Amazon (Brazil)& MODIS & 480m - 3 obs. (months) & HRM (Landsat-derived) & 2 & No & STSU\\ \hline

\multirow{3}{4em}{\cite{wang2021real} 2021} & Daxing (China)& MODIS & 480m - 2 obs. (months) & HRM (Landsat-derived) & 2 & No & RSTSU \\
 & Zibo (China)& MODIS & 480m - 2 obs. (months) & HRM (Landsat-derived) & 2 & No & RSTSU\\ 
 & Lichuan (China)& MODIS & 480m - 2 obs. (months) & HRM (Landsat-derived) & 2 & No & RSTSU\\ \hline

\multirow{2}{4em}{Ours} & Andalusia (Spain)& MODIS & 460m - 12 obs. (months) & HRM (SIPNA L1) & 4 & Yes & LSTM-based \\ 
 & Andalusia (Spain)& MODIS & 460m - 12 obs. (months) & HRM (SIPNA L2) & 10 & Yes & LSTM-based \\ 
 
\end{tabular}}

\label{table:2}
\end{table*}
\renewcommand{\arraystretch}{1}

\subsection{Labeled datasets based on LULC products}

Supervised learning requires high amounts of ground truth data to achieve better generalization. One of the biggest limitations in spectral unmixing is the limited availability of ground truth LULC maps \cite{9324546,zhu2017hyperspectral}. Some areas or regions, especially in western countries, have LULC ground truth based on visual interpretation for specific fields of study. For example, SIPNA \cite{SIPNA} was intended for territorial planing in Spain. Its annotation was carried out by experts during several years. This dataset can be used to annotate RS data.

In parallel, there exist several annotated MS multemporal datasets  prepared for supervised spectral unmixing (see Table \ref{table:2}). However, all of them are private.

Our work is the first in providing a public good quality MS multitemporal mixed pixel labeled dataset, named Andalusia-MSMTU, that includes not only spectro-temporal information but also geo-topographic and climatic ancillary data. Andalucía-MSMTU is organized into two hierarchical levels of classes with four and ten LULC types and it is especially suitable for building umimixing DL-based models for LULC abudance estimation. The proposed methodology constitutes the state-of-the art in Andalusia-MSMTU.

\section{Preliminaries and background} \label{sec:back}
We define a multivariate time series as a sequence of observations $X=(\mathbf{x}_{1}, \mathbf{x}_{2}, ..., \mathbf{x}_{T})$, where $T$ is the number of observations or time steps. Each observation $\mathbf{x}_{t} \in \mathbb{R}^{C}$ where $t \in \{1, ..., T\}$ consists of $C$ variables, such that $\mathbf{x}_{t} = \{x_{t}^{1}, x_{t}^{2}, ..., x_{t}^{C}\}$.

\subsection{Recurrent Neural Network}

RNN \cite{rumelhart1985learning} is a NN architecture specifically designed for handling sequential data. RNN consider the sequential relationship of inputs by using a shared function $f$ to process each input. RNN process the time series using a recurrence approach at every time step $t$, computing a hidden state $\mathbf{h}_{t}$ by considering the previous hidden state $\mathbf{h}_{t-1}$ and the current input $\mathbf{x}_{t}$:

\begin{equation}\label{eq:rnn_hidden}
        \mathbf{h}_{t} = f(\mathbf{h}_{t-1}, \mathbf{x}_{t})
\end{equation}

where $\mathbf{h}_{0}$ is normally, at the beginning, the zero vector, i.e., $\mathbf{h}_{0} = \mathbf{0}$.

There are several choices on how to process sequential information. In this work, we focus on the LSTM network, which is an improvement of normal RNN solving some of its biggest limitations \cite{hochreiter1997long}

\subsection{BRITS}
In time series data and specifically in RS data, it is common to find missing values due to sensor errors, cloud cover and more \cite{gerber2018predicting}. To handle this situation, there exists a type of RNN architecture that can learn to solve two tasks simultaneously: imputing missing values and classifying the input sequence data. This model is called Recurrent Imputation for time series (RITS) \cite{cao2018brits}. The RITS model perform the imputation algorithm to assist the classification task and obtain the final classification as:

\begin{equation}
        \mathbf{\hat{y}} = f_{out}(\mathbf{h}_{T}) 
\end{equation}

where $\mathbf{\hat{y}}$ is the final classification, $f_{out}$ is the classification function, and $\mathbf{h}_{T}$ is the last hidden state.

\begin{figure*}[h!]
        \centering
    	\includegraphics[width=0.75\textwidth]{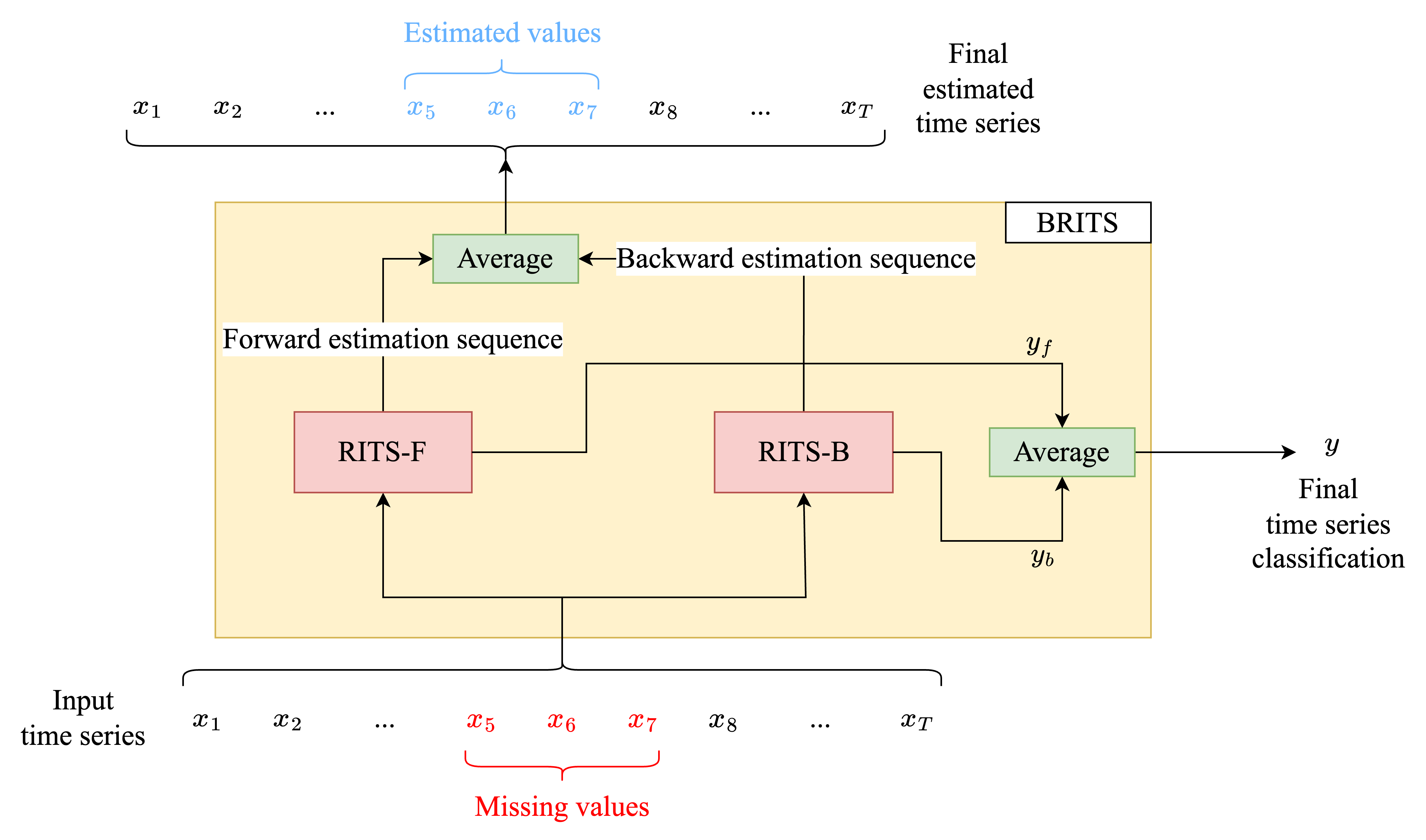}
    	\caption{BRITS architecture.}\
    	\label{fig:BRITS}
\end{figure*}

In practice, considering only unidirectional forward dynamic is problematic due to slow convergence, inefficiency in training and bias exploding problem \cite{cao2018brits}. To overcome these issues a bidirectional version named Bidirectional RITS (BRITS) model is proposed also in \cite{gerber2018predicting} to learn forward and backward patterns by accessing information from past and future at any given time step. The final scheme of BRITS can be seen in Figure \ref{fig:BRITS}.

\section{Study area and data construction} \label{sec:study}
This section describes the study area and provides full details on how the used dataset was built and processed. 

\subsection{Study area}
\begin{figure*}[h!]
    \centering
    \includegraphics[width=0.8\textwidth]{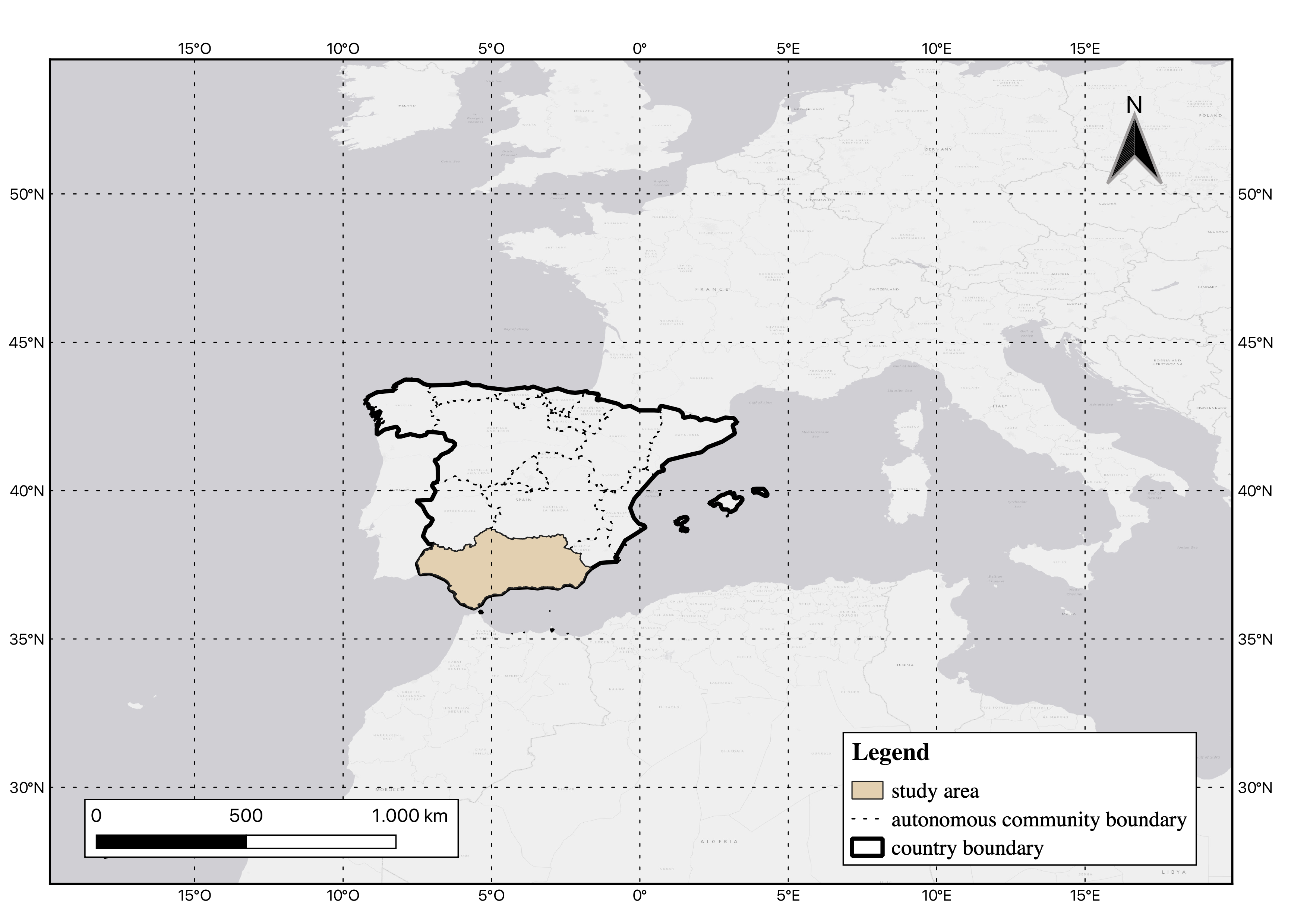}
    \caption{Study area: Andalusia, Spain.}\
    \label{fig:study_area}
\end{figure*}

Andalusia is the second-largest, most populous, and southernmost autonomous community in Peninsular Spain (Figure \ref{fig:study_area}). Andalusia is one of the most biodiverse and heterogeneous regions of Europe. It contains steep altitudinal gradients, and it has a wide variety of landscapes and climatic conditions which results in a vast variety of vegetation types that hold the greatest diversity of plant and animal species in Europe. The long and dynamic history of human activities has also led to a complex landscape configuration with frequent mosaics of small patches of different types of natural, semi-natural land covers and human land uses. Hence, Andalusia offers an ideal field laboratory to test the creation of detailed and fine scale LULC maps containing the abundance of each LULC class per pixel to monitor the socioeconomic and environmental dynamics in complex landscapes using DL and MS time series of satellite imagery.

\subsection{Andalusia-MSMTU dataset construction} \label{sec:dataset}
To build Andalusia MultiSpectral, MultiTemporal Unmixing dataset, named Andalusia-MSMTU, several sources were utilized: MODIS, SRTM  digital elevation data \cite{farr2007shuttle}, REDIAM's environmental information \cite{REDIAM} and SIPNA \cite{SIPNA}. Herein, three different processes were used to create the dataset, (1) MODIS MS time series extraction, (2) ancillary data extraction and LULC abundances annotations.

\subsubsection{MODIS pixel time series extraction} \label{sec:data_extraction}

The time series data were extracted from two satellites Terra and Aqua using MODIS sensor at 460m spatial resolution and at monthly temporal resolution. As LULC changes during one year are very limited in a $460m$ pixel, we assume that the LULC abundances are representative of the full year.

Spatio-temporal filtering was applied using MODIS ‘Quality Assessment’ (QA) flags and the "State QA" flags. Moreover, as the process of Terra and Aqua data filtering generates many missing values, to further reduce the amount of noise in the data, two solutions were employed: (1) the 8-days time series data were transformed into monthly composites by computing the monthly mean from the individual observations, then (2) the monthly data from the Terra and Aqua satellites were combined to generate a merged Terra+Aqua monthly dataset. All this process was performed in Google Earth Engine (GEE) \cite{gorelick2017google} and inspired by \cite{khaldi2022timespec4lulc}.

\subsubsection{Ancillary data extraction for each MODIS pixel}

In addition to the MODIS data, for every pixel we included geographic, topographic, and climatic ancillary information. Pixel longitude and latitude were directly extracted from MODIS metadata. Pixel altitude was obtained using the SRTM 30m/pixel digital elevation model \cite{farr2007shuttle}. MODIS pixel slopes were calculated using GEE slope calculation algorithm on the same 30m elevation model. Finally, climatic data were downloaded from REDIAM's environmental information \cite{REDIAM}, including potential evapotranspiration, precipitation, mean annual temperature, mean of the maximum temperatures, and mean of the minimum temperatures. All types of ancillary data came in different resolutions or scale, so to match the resolution of our MODIS pixels we computed the average across all finer resolution pixels inside each MODIS pixels to obtain the value at 460m resolution.

\subsubsection{Pixels' LULC abundances annotation from SIPNA}

\begin{figure*}[h!]
    \centering
    \includegraphics[width=\textwidth]{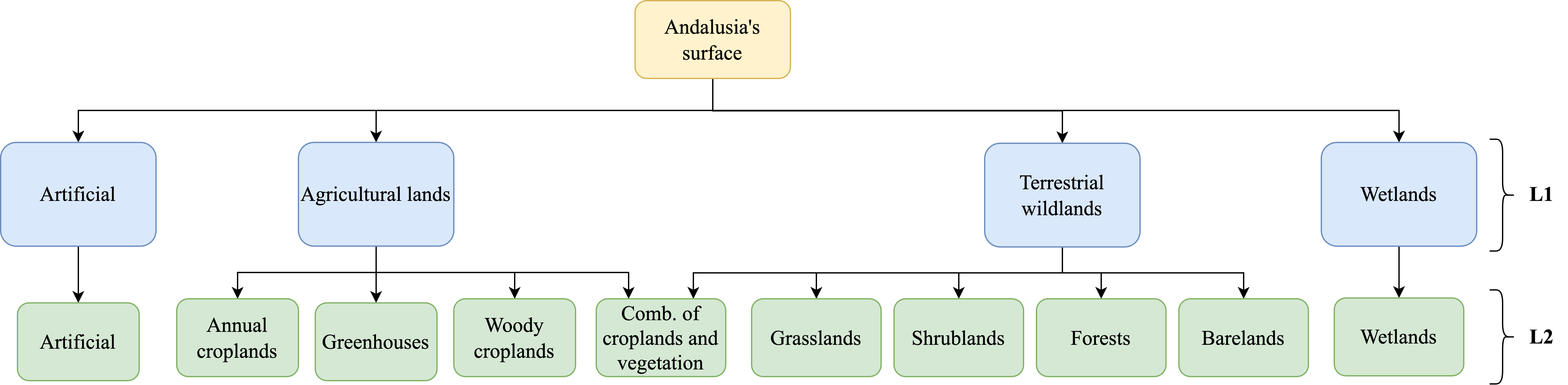}
    \caption{Hierarchical structure of the SIPNA-based LULC classes. The blue boxes represent the level 1 (L1) classes. The green boxes represents the Level 2 (L2) classes.}\
    \label{fig:class_hierarchy}
\end{figure*}

To annotate each $460m$ MODIS pixel with the abundance of each LULC class, the official LULC map of Andalusia for the year 2013 (SIPNA) \cite{SIPNA} was used. Given the coarse resolution of MODIS pixels we only considered level 1 (four classes) and an adapted version of level 2 (ten classes) of the classification hierarchy of SIPNA (Figure \ref{fig:class_hierarchy}).

Given that SIPNA provides information at sub-pixel level, we calculated the exact abundances of all the LULC classes existing in each  MODIS $460m$ resolution pixel, as illustrated in Figure \ref{fig:flowchart_props}, using QGIS software \cite{QGIS_software} as follows: the SIPNA polygons were first converted to raster format providing a LULC map at $10m$ resolution. The rasterized map was then converted to match the spatial resolution of MODIS by counting the number of $10m$ resolution pixels of each LULC class and dividing them by the total number of $10m$ resolution pixels inside each $460m$ resolution, resulting in the class proportions for each $460m$ pixel of Andalusia. Finally, the MODIS pixels abundances annotations were coupled with their corresponding time series and ancillary data to obtain the Andalusia-MSMTU dataset. With the help of several RS expert, we visually assess that the $10m$ resolution was suitable for the rasterization. The proposed values were $100m$, $50m$, $10m$ and $5m$. The $100m$ and $50m$ resolution pixels were too coarse to maintain the quality of the different polygon annotations. The $10m$ and $5m$ resolution pixels were great to maintain the information and we finally decided to rasterize the polygons to $10m$ resolution because of computational and time convenience, since the $5m$ raster was $4$ times more expensive than the $10m$ raster.

\begin{figure*}
\centering
\begin{subfigure}{0.45\textwidth}
    \includegraphics[width=\textwidth]{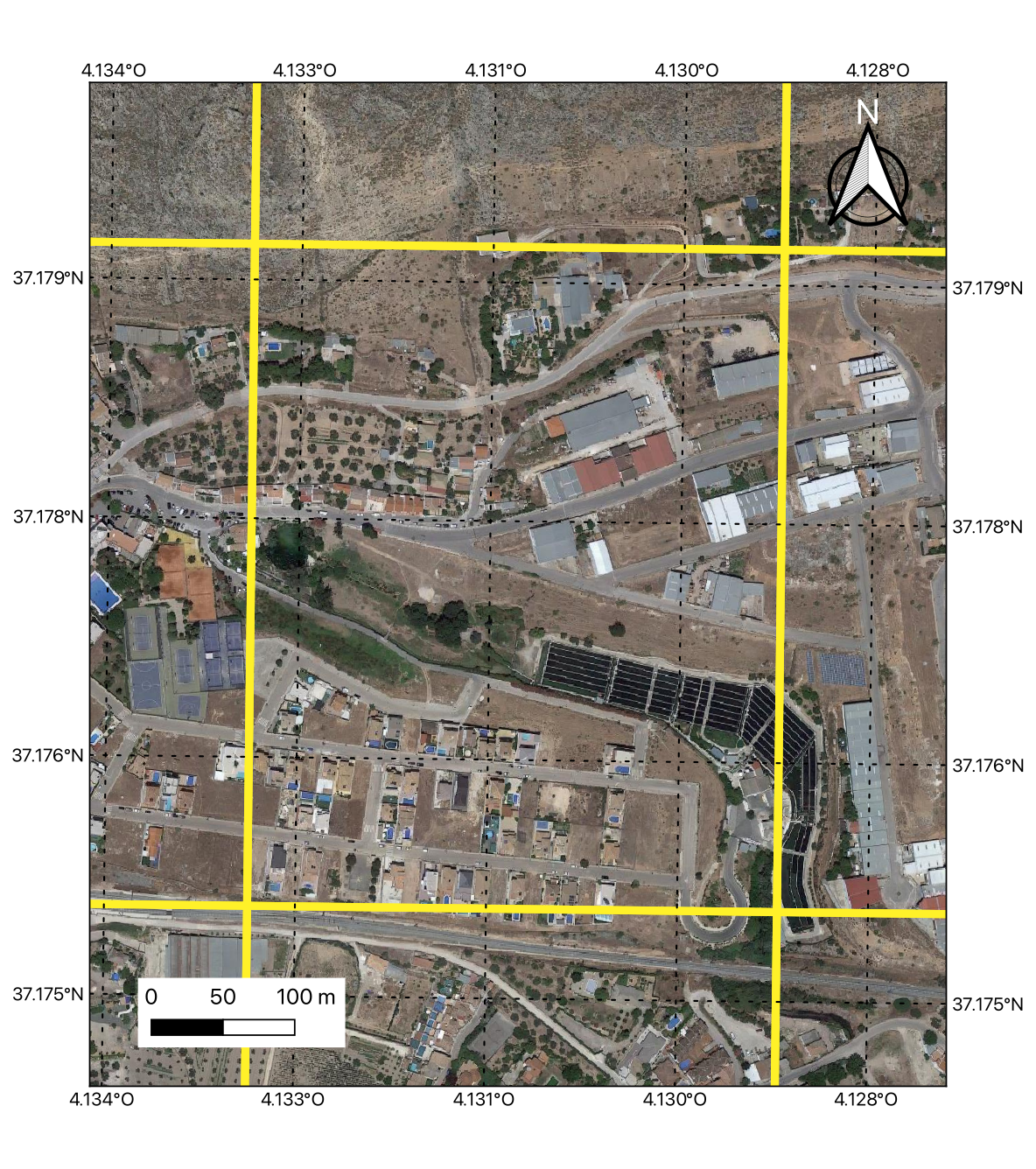}
    \caption{Google satellite image of one MODIS pixel.}
    \label{fig:first}
\end{subfigure}
\hspace{1cm}
\begin{subfigure}{0.45\textwidth}
    \includegraphics[width=\textwidth]{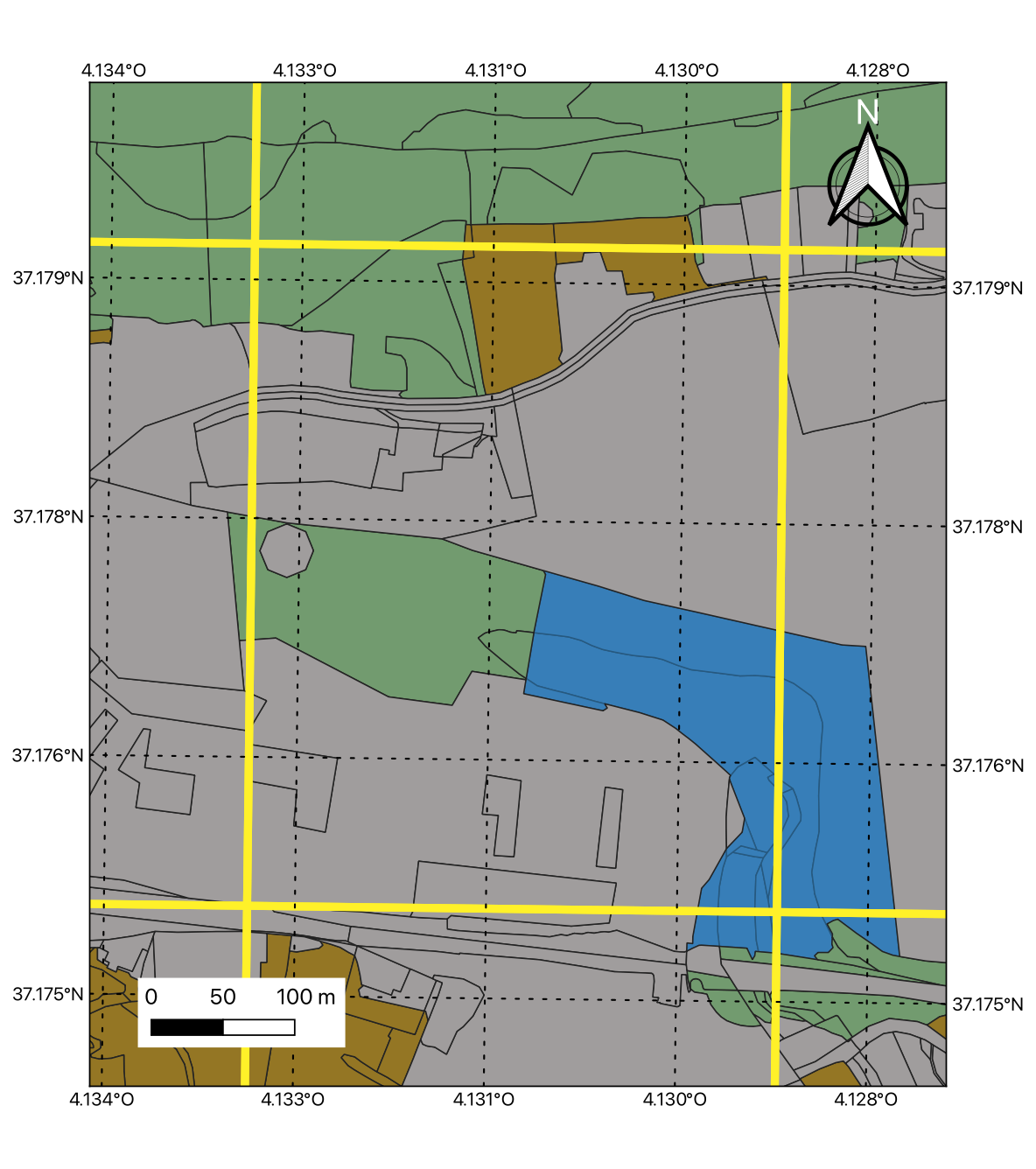}
    \caption{SIPNA LULC map (polygon format).}
    \label{fig:second}
\end{subfigure}
\hspace{1cm}
\begin{subfigure}{0.45\textwidth}
    \includegraphics[width=\textwidth]{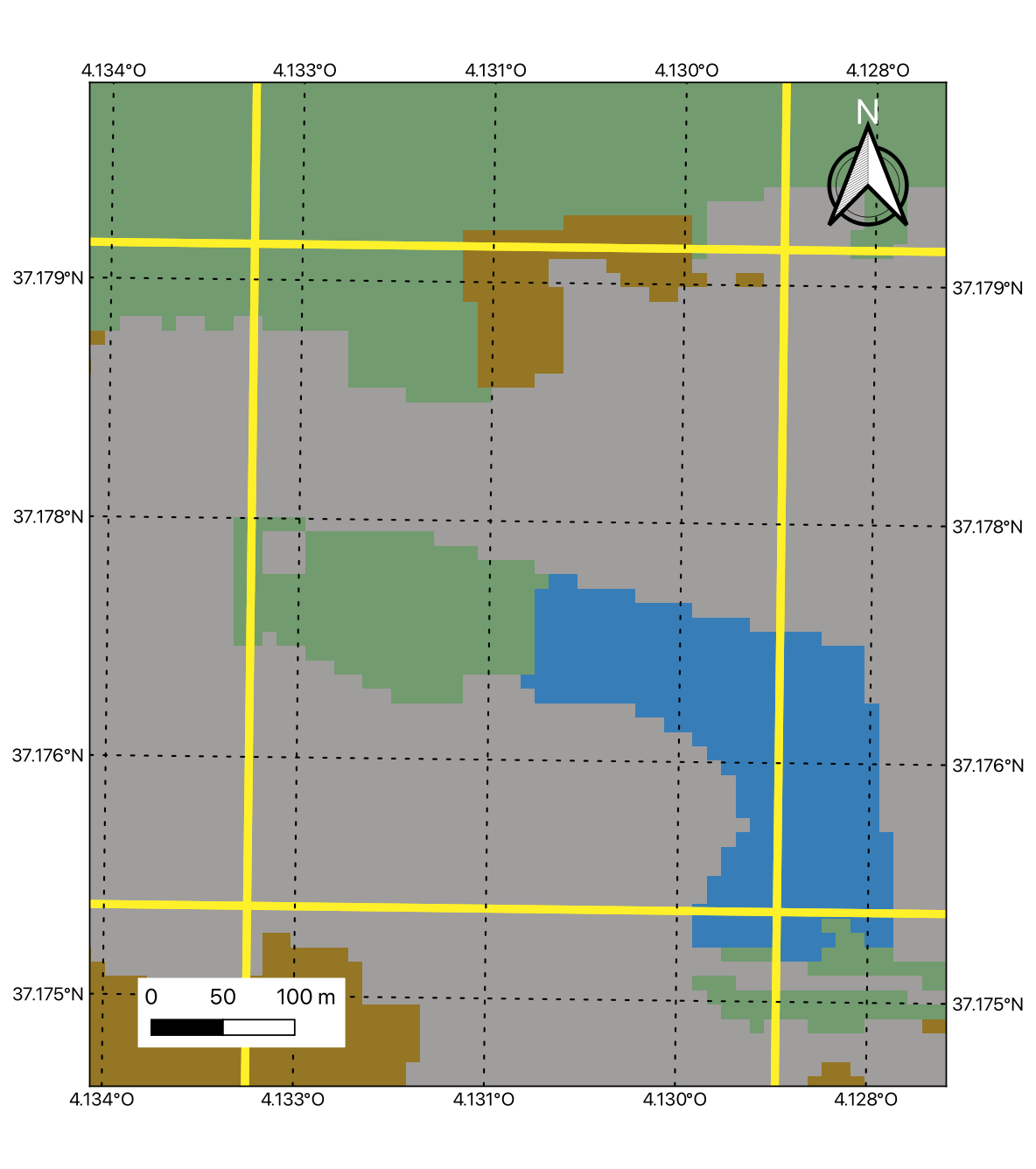}
    \caption{SIPNA LULC map ($10m$ raster format).}
    \label{fig:third}
\end{subfigure}
\hspace{1cm}
\begin{subfigure}{0.45\textwidth}
    \includegraphics[width=\textwidth]{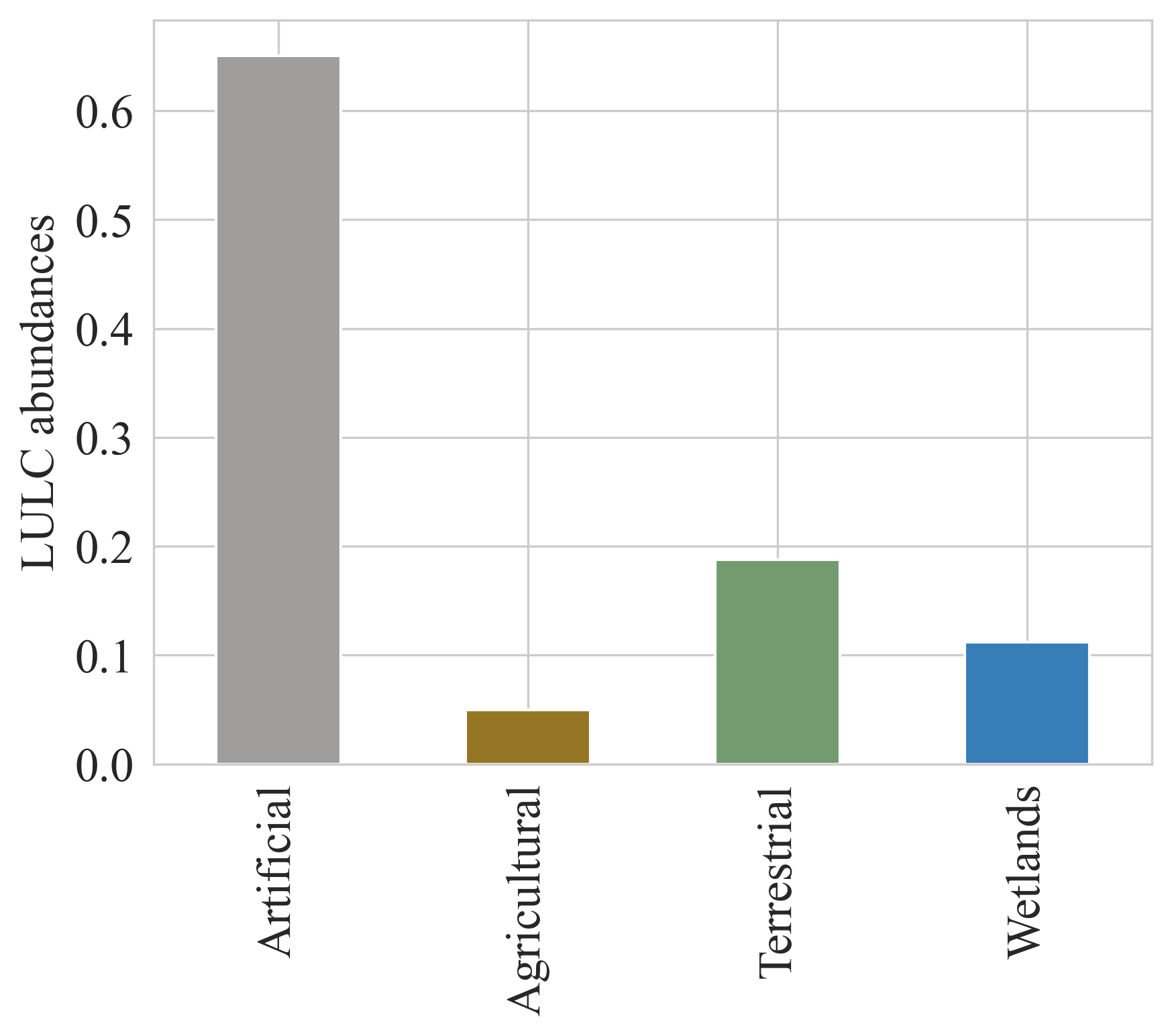}
    \caption{Level 1 LULC class abundances.}
    \label{fig:fourth}
\end{subfigure}

\vspace*{+2mm}
\caption{Example of how class abundances are obtained for each MODIS pixel. (a) shows the satellite image of the Google Satellite corresponding to one MODIS pixel. (b) shows the annotated SIPNA polygons. (c) the rasterized LULC map at 10m resolution. (d) the obtained abundance of level 1 classes for that MODIS pixel.}
\label{fig:pixel}
\end{figure*}

\begin{figure*}[]
\centering
\begin{subfigure}{0.4\textwidth}
    \includegraphics[width=\textwidth]{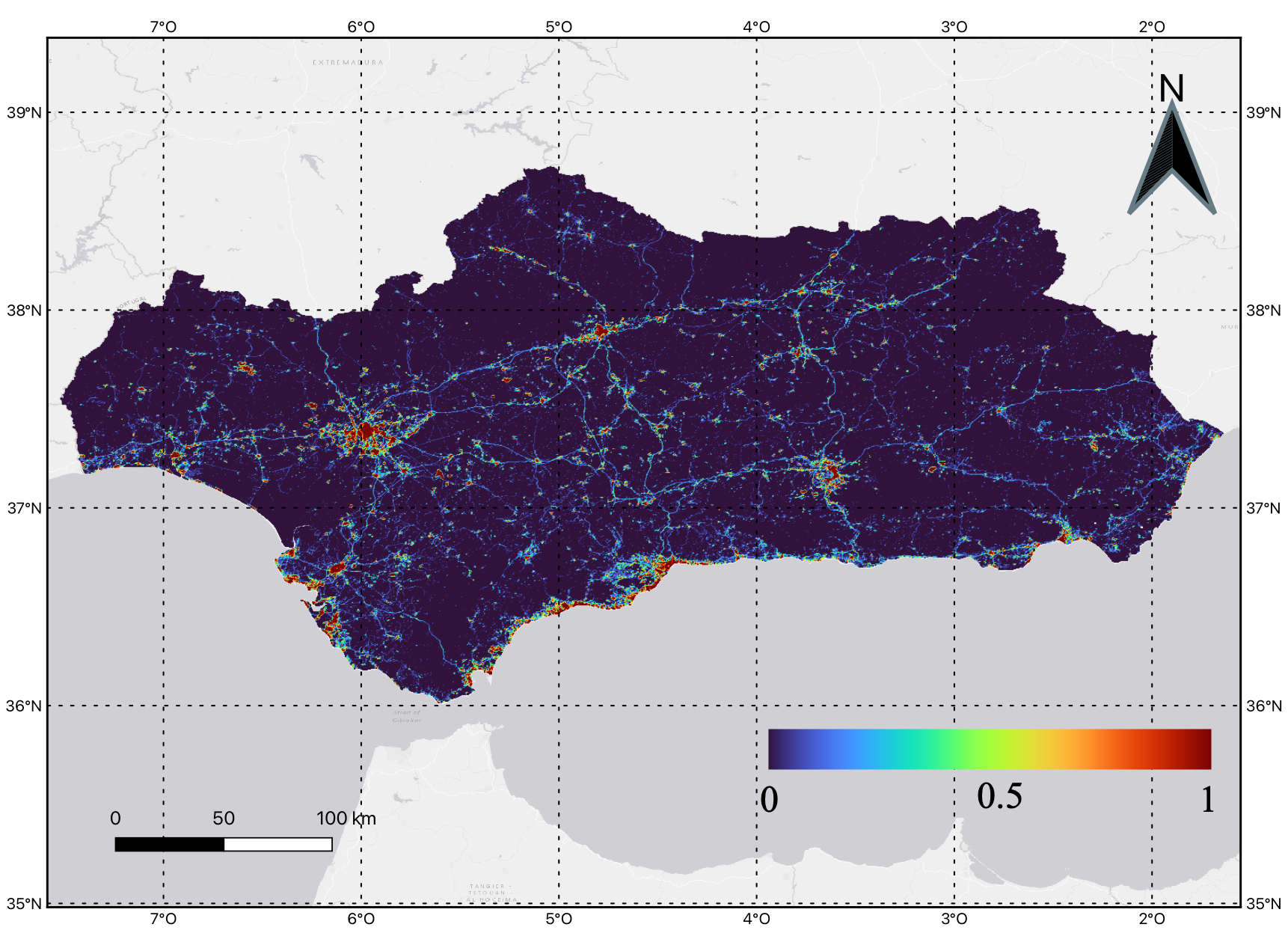}
    \caption{Artificial}
    \label{fig:first}
\end{subfigure}
\hspace{1cm}
\begin{subfigure}{0.4\textwidth}
    \includegraphics[width=\textwidth]{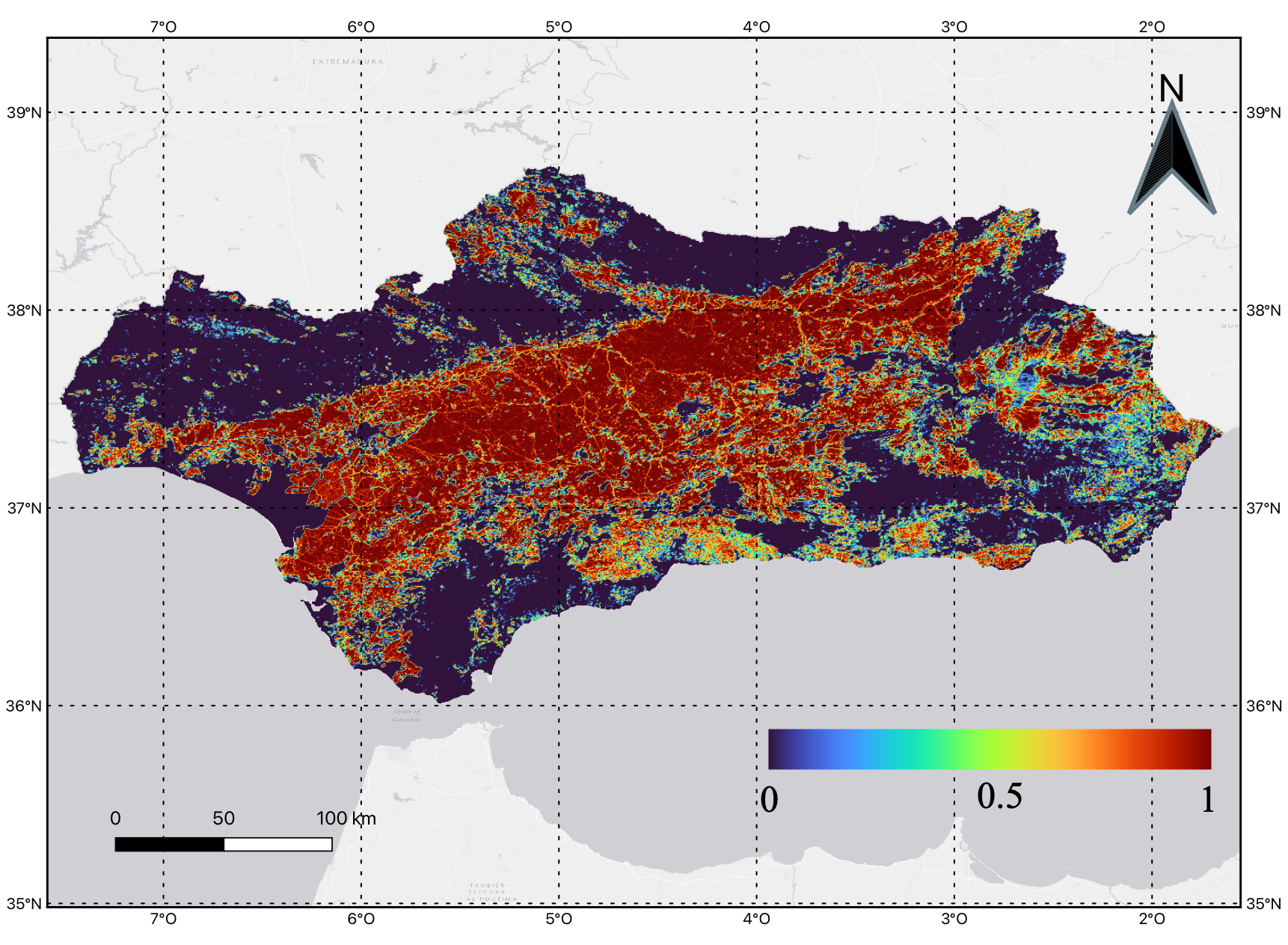}
    \caption{Agricultural lands}
    \label{fig:second}
\end{subfigure}
\hspace{1cm}
\begin{subfigure}{0.4\textwidth}
    \includegraphics[width=\textwidth]{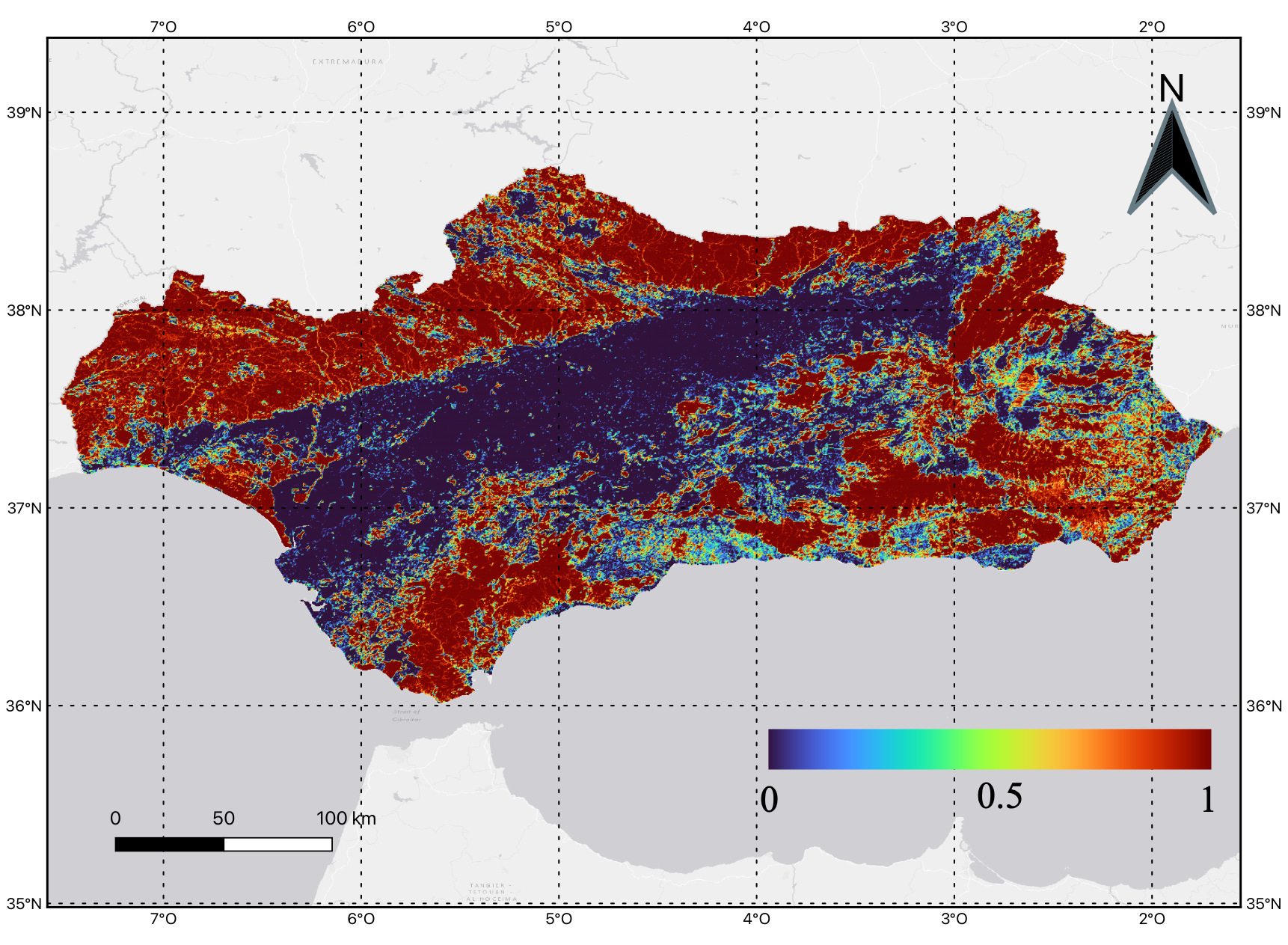}
    \caption{Terrestrial lands}
    \label{fig:third}
\end{subfigure}
\hspace{1cm}
\begin{subfigure}{0.4\textwidth}
    \includegraphics[width=\textwidth]{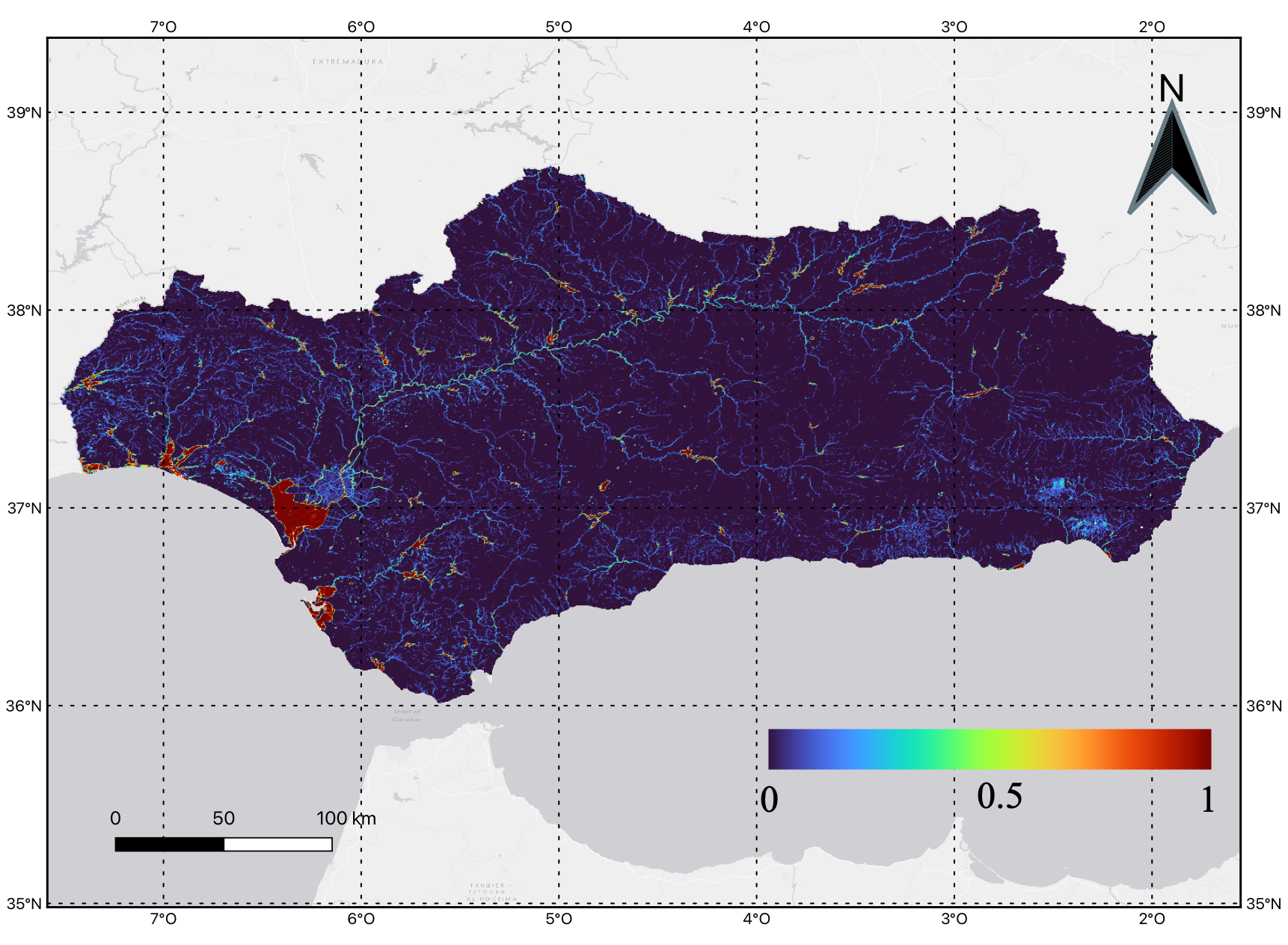}
    \caption{Wetlands}
    \label{fig:fourth}
\end{subfigure}
\vspace*{+2mm}        
\caption{Gradient map of abundances from Andalusia 460m pixels for each LULC class in level 1 of SIPNA: (a) Artificial, (b) Agricultural lands, (c) Terrestrial lands, and (d) Wetlands. Blue pixels represent low abundance and red pixels represent high abundance.}
\label{fig:proportions}
\end{figure*}

In Figure \ref{fig:pixel} it is showed an example of the calculation of class abundances for a given pixel. An illustrative example of the distribution of abundances of LULC classes in level 1 of the classification hierarchy over the Andalusia territory is displayed in Figure \ref{fig:proportions}, being "agricultural lands" and "terrestrial lands" the classes that dominate the Andalusian territory. Andalusia-MSMTU dataset \cite{rodriguez_ortega_jose_2023_7752348} is available in a public data repository hosted by \href{https://zenodo.org/}{Zenodo} at: \url{https://zenodo.org/records/7752348}

\begin{figure*}[h!]
    \centering
    \includegraphics[width=\textwidth]{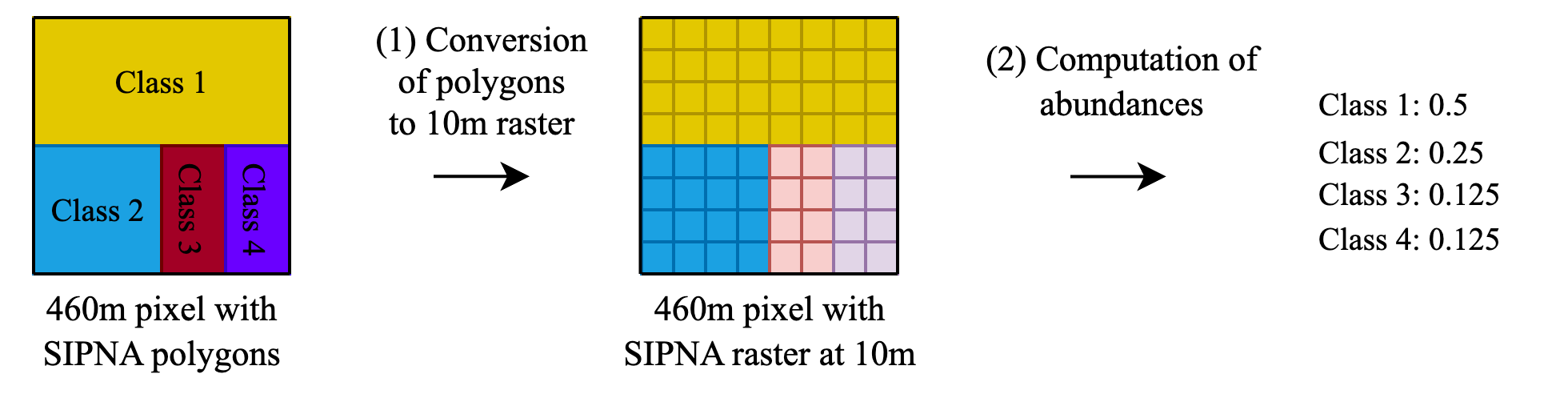}
    \caption{The used scheme for extracting class abundances in every MODIS pixel of Andalusia. (1) The original SIPNA polygons were converted to a $10m$ raster, then (2) the LULC abundances were computed for each $460m$ pixel.}\
    \label{fig:flowchart_props}
\end{figure*}

\section{Methodology} \label{sec:metho}
Given the success of DL methods in learning tasks, we propose an RNN-based model to learn the input-output relationship between the remotely sensed MS time series and the corresponding LULC abundances in a pixel. We do this by building a high quality annotated dataset as described in Section \ref{sec:dataset} and splitting it in training and test sets. We use the training dataset to train the DL model in a supervised manner. The test set is composed of unseen samples during the training and is used to evaluate the true performance of the model.

Formally, we have a set of $n$ MS time series pixels $\{X_{1}, X_{2}, ..., X_{n}\}$  with their corresponding class abundances $\{y_{1}, y_{2}, ..., y_{n}\}$ where $y_i \in S^{C}, i \in [1,n]$. $S^{C}$ is the sample space of class abundances commonly referred to as the simplex \cite{aitchison1982statistical}. In our case $C$ is equal to 4 and 10 for level 1 and 2 of the hierarchy, respectively. 

To enhance class abundance estimation further, in addition to using the MS multitemporal data we also include ancillary information from two types:

\begin{itemize}
    \item Geo-topographic data: Geographical coordinates (longitude and latitude), altitude and slope. Incorporating geographic coordinates can help the model understand the spatial distribution of land cover types, which can be valuable in guiding the spectral unmixing process and making it more contextually accurate. Similarly, adding topographic data (altitude and slope) provides useful information that complements the spectral characteristics of a pixel. In fact, terrain slope is known to influence surface reflectance, so incorporating it into the model can allow slope-related changes in reflectance to be taken into account, making its predictions more robust.
    
    \item Climatic data: Precipitation, potential evapotranspiration, mean temperature, maximum temperature and minimum temperature.  Some land cover classes, such as agricultural lands, forests, and wetlands, respond differently to variations in climate. By using climatic variables, the DL model can distinguish between these climate-dependent classes more effectively.
\end{itemize}

Below, we describe the architecture of the used model and the evaluation metrics.

\subsection{Model architecture}

\begin{figure*}[h!]
    \centering
	\includegraphics[width=0.85\textwidth]{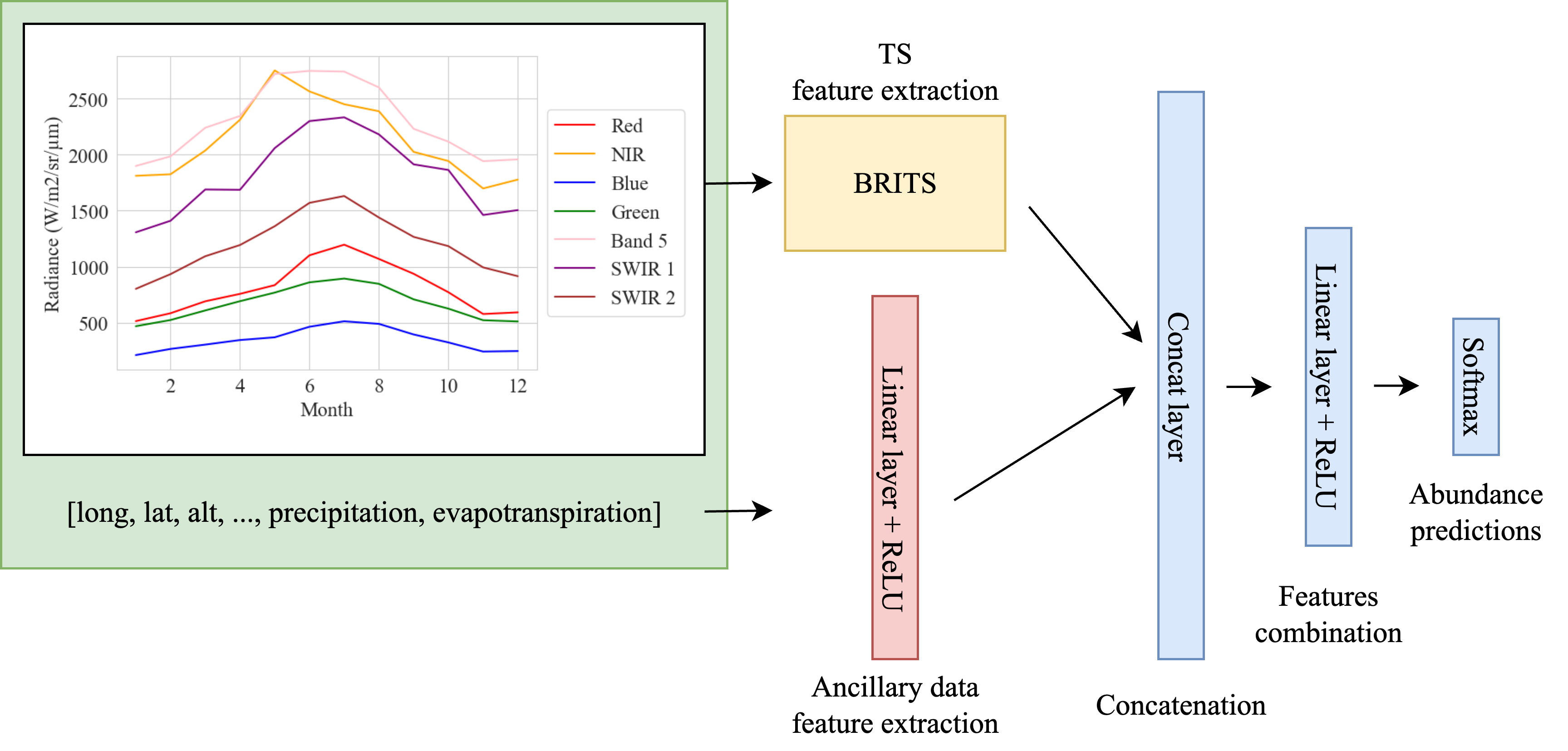}
	\caption{Our proposed neural network. The green box denotes the input data for a given pixel, i.e., MS time series data + ancillary data. The yellow box denotes the BRITS model for MS time series feature extraction, the red box denotes the ancillary data feature extraction layer and the blue boxes denote the final layers for features combination and softmax transformation of neural network's output.}
	\label{fig:metho}
\end{figure*}

Our BRITS-based approach to estimate the class abundances using  MS multitemporal data and ancillary information for each mixed pixel is depicted in Figure \ref{fig:metho}. 
The proposed approach includes three components: 

\begin{enumerate}
    \item Spectro-temporal feature extraction:  We use BRITS model \cite{cao2018brits} to extract the spectro-temporal patterns in presence of missing values from our dataset. 
    
    \item Ancillary data feature extraction: To incorporate ancillary information, we process the external information using a linear layer with ReLU non-linearity. 
    
    \item Concatenation and features combination: The output features of part (1) and (2) are concatenated and processed by a final dense layer that outputs  $C$ (the number of classes) scores.
\end{enumerate}

The final dense layer generates an unbounded outputs $\mathbf{o}$ where $\mathbf{o} \in \mathbb{R}^{C}$ with $C$ being the number of classes. Following the work of \cite{heremans2016effect}, we applied the softmax transformation to obtain the final abundances predictions $\mathbf{a} \in S^{C}$:

\begin{equation}
    a_{j} = \frac{e^{o_{j}}}{\sum_{c=1}^{C} e^{o_{c}}}
\end{equation}

where $a_{j}$ denotes the abundance prediction for the $j$th class, $o_{j}$ denotes the final layer's output associated with the $j$th class, and $e$ denotes the exponential function.

Finally, the NN is optimized by minimizing the mean-square error ($MSE$) between the abundances predictions and the reference abundances:

\begin{equation}
    MSE = \frac{\sum_{i=1}^{N} \sum_{c=1}^{C} (r_{ic} - a_{ic})^{2}}{N}
\end{equation}

where $r_{ic}$ and $a_{ic}$ are the reference abundance and the predicted abundance, respectively, for the $c$th class in the $i$th sample, and $N$ is the number of training samples.

\subsection{Evaluation criteria}
To assess the effectiveness of the proposed unmixing model, four regression metrics are examined:

\begin{itemize}
    \item Pearson's Correlation Coefficient (CC):
    \begin{equation}
        CC = \frac{\sum_{i=1}^{N} (r_{i} - \overline{r})(a_{i} - \overline{a})}{\sqrt{\sum_{i=1}^{N} (r_{i} - \overline{r})^{2} \sum_{i=1}^{N} (a_{i} - \overline{a})^{2}}}
    \end{equation}

    \item Root Mean Squared Error (RMSE):
    \begin{equation}
        RMSE = \sqrt{\frac{1}{n} \sum_{i=1}^{N} (r_{i} - a_{i})^{2}}
    \end{equation}

    \item Relative Root Mean Squared Error (RRMSE):
    \begin{equation}
        RRMSE = \sqrt{\frac{\sum_{i=1}^{N} (r_{i} - a_{i})^{2}}{\sum_{i=1}^{N} (r_{i})^{2}}}
    \end{equation}

    \item Mean Absolute Error (MAE):
    \begin{equation}
        MAE = \frac{1}{n} \sum_{i=1}^{N} |r_{i} - a_{i}|
    \end{equation}
\end{itemize}

where $r_{i}$ is the reference abundance, $a_{i}$ the predicted abundance, and $\overline{r}$ and $\overline{a}$ are the mean of both variables. Finally, we also considered F1-score (Formula \ref{eq:f1}) to evaluate how good is the model in predicting the majoritarian class in each mixed pixel.

\begin{equation} \label{eq:f1}
        F1 = \frac{2*TP}{2*TP+FP+FN}
\end{equation}

\begin{figure}[h]
    \centering
    \includegraphics[width=\columnwidth]{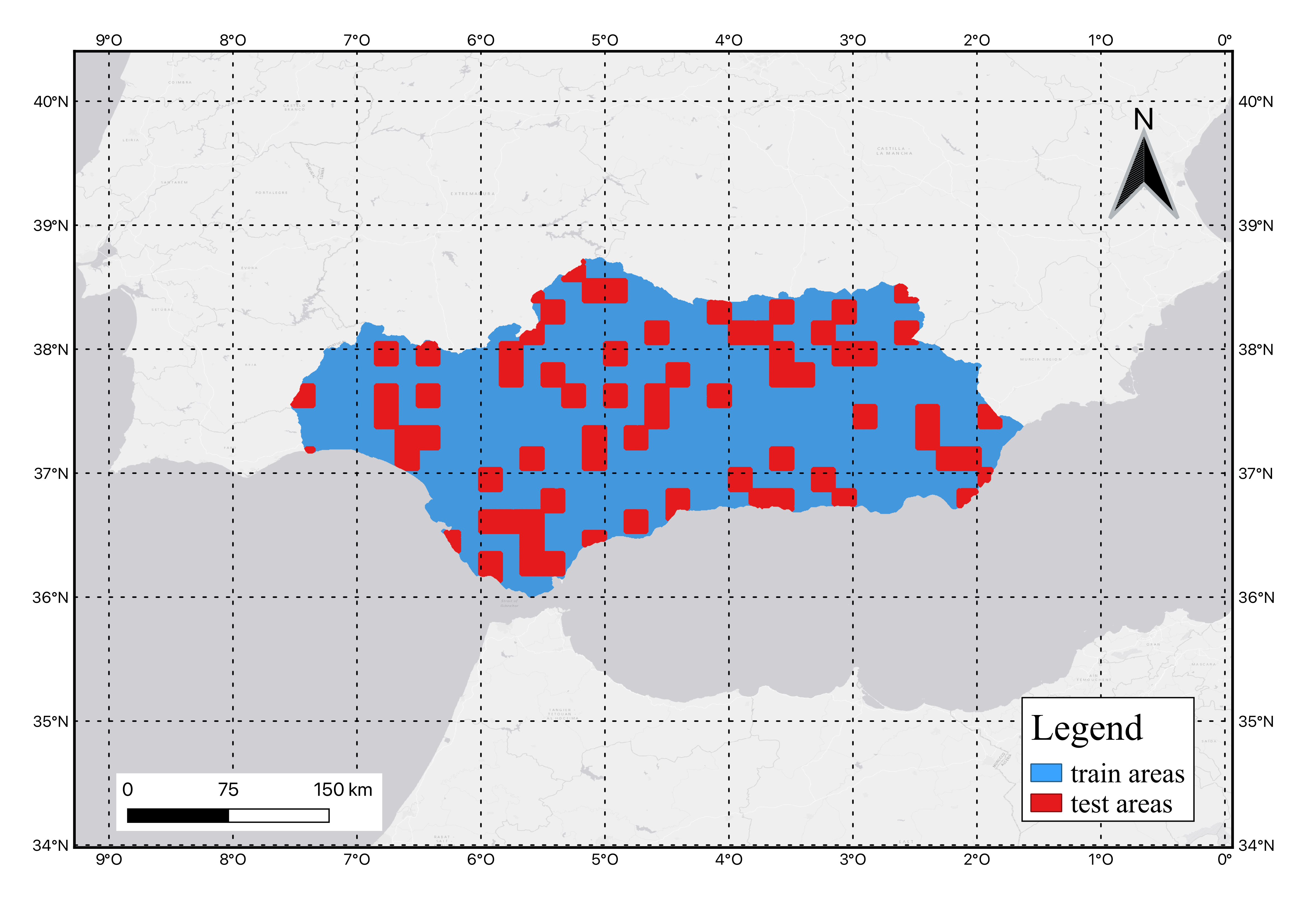}
    \caption{Train (blue) and test (red) areas.}\
    \label{fig:block_split}
\end{figure}
\subsection{Experimental design}

To analyse the effect of introducing ancillary data and using different levels of the LULC legend on the performance of our DL approach for spectro-temporal unmixing, we considered different input data combinations, that is: using (1) only the MS time series data, (2) time series plus geo-topographic ancillary data, (3) time series plus climatic ancillary data, and (4) time series plus geo-topographic and climatic ancillary data.

In order to avoid spatial autocorrelation of neighbouring pixels we used a block train test splitting \cite{roberts2017cross, uieda2018}. Firstly, we divided the entire Andalusian territory in areas of equal size using blocks of 18x15 kilometers, which means that each block contains 1250 of 460m pixels approximately. Subsequently, 80\% of the pixel blocks were assigned randomly to the training set, with the remaining 20\% allocated to the test set. Figure \ref{fig:block_split} illustrates the areas of pixels designated for the training and testing sets.  The source code to run these experiments will be available after acceptance at \url{https://github.com/jrodriguezortega/MSMTU}.

\textbf{Implementation details:}
Our models undergo training using the Adam optimizer \cite{kingma2014adam} for a total of 200 epochs with a batch size of 2048. We initialize the learning rate at 0.003 and progressively reduce it via the cosine learning rate decay scheduler. All experiments were conducted utilizing the PyTorch deep learning framework \cite{paszke2019pytorch}.

\textbf{TimeSpec4LULC \cite{khaldi2022timespec4lulc} pre-training:}
TimeSpec4LULC is an open-source dataset comprising MS time series data for 29 LULC classes, designed for training machine learning models. This dataset is constructed using the seven spectral bands from MODIS sensors, providing data at a 460m resolution, spanning the time period from 2000 to 2021. We found that pre-training BRITS model on TimeSpec4LULC dataset and fine-tuning it on Andalusia-MSMTU  provides better results that training it from scratch, mainly because of the similarity between both datasets.

\section{Experimental results} \label{sec:results}
 This section provides the experimental results of the proposed model at SIPNA level 1 and level 2. 
\subsection{SIPNA level 1}

We evaluated the proposed model in Section \ref{sec:metho} on different combinations of  spectro-temporal data and ancillary data. In particular, we considered these combinations: (spectro-temporal data), (spectro-temporal data + geo-topographic data),  (spectro-temporal data + climatic data) and (spectro-temporal data + geo-topographic and climatic data). Besides, we also include the results of a a baseline model trained from scratch on spectro-temporal data only to show how the pretraining on TimeSpec4LULC dataset is highly beneficial.
The results of these five models in terms of the average MAE, RMSE, RRMSE, CC, F1-score, RRMSE gain, CC gain across the four classes of level 1 are provided in Table \ref{table:results_n1}. Aditionally, the computational complexity of each model is expressed in terms of MFLOPs in the last column. Firstly, we can see in the first two rows that by just pretraining our model in TimeSpec4LULC dataset improves the results in every metric, proving the value of pretraining DL models in similar tasks to achieve better performance. Secondly, it can be seen that including ancillary information always improves the spectral unmixing performance with respect to the baseline model (using MS time series only and trained from scratch). 
The highest performance is achieved when including both, geo-topographic and climatic data together with the MS time series showing the lowest MAE, RMSE and RRMSE, with 1.10\%, 1.17\%  and  3.39\% of improvement respectively and highest CC and F1-score with 0.0276 and 0.0216 of improvement respectively, with respect to the baseline model.


\renewcommand{\arraystretch}{1.5}
 \begin{table*}[h!]
\centering
\caption{Performance comparison of our model trained from scratch (first row) and finetuned from TimeSpec4LULC (second row) using only multi-spectral multi-temporal input data, by adding  geo-topographic data only (third row), by adding climatic data only (fourth row) and by adding both geo-topographic and climatic (fifth row). The performance is expressed in terms of average Mean Absolute Error (MAE), Root Mean Squared Error (RMSE), Relative Root Mean Squared Error (RRMSE), Correlation Coefficient (CC), F1-score, RMSE gain and CC gain with respect to baseline model for SIPNA level 1 classes. Last column, "MFLOPs", indicates the model's computational complexity in terms of Mega FLOPs.}
\vspace{+1mm}
\resizebox{0.98\textwidth}{!}{%
\begin{tabular}{l|r r r r r r r | r}
 \textbf{Ancillary input data} & \textbf{MAE (\%)} & \textbf{RMSE (\%)} & \textbf{RRMSE (\%)} & \textbf{CC} & \textbf{F1-score} & \textbf{RMSE (\%) gain} & \textbf{CC gain} & \textbf{MFLOPs} \\ \hline
  None (from scratch) & 7.25\% & 12.82\% & 43.73\% & 0.8447 & 0.8055 & 0.0000\% & 0.0000 &  2.3055 \\
 None (finetuned from TimeSpec4LULC) & 6.70\% & 12.28\% & 42.40\% & 0.8570 & 0.8140 & -0.5400\% & 0.0123 & 2.3055 \\
 Climatic & 6.55\% & 12.24\% & 42.34\% & 0.8582 & 0.8107 & -0.5800\% & 0.0243 &  2.3665 \\
 Geo-topographic & 6.21\% & 11.80\% & 40.84\% & 0.8691 & 0.8195 & -1.0200\% & 0.0244 & 2.3663 \\ 
 Geo-topographic + climatic & \textbf{6.15\%} & \textbf{11.65\%} & \textbf{40.34\%} & \textbf{0.8723} & \textbf{0.8241} & \textbf{-1.1700\%} & \textbf{0.0276} & 2.3673 \\ 
\end{tabular}
}
\label{table:results_n1}
\end{table*}
\renewcommand{\arraystretch}{1} 

\begin{figure*}[h!]
\centering
\begin{subfigure}{0.45\textwidth}
    \includegraphics[width=\textwidth]{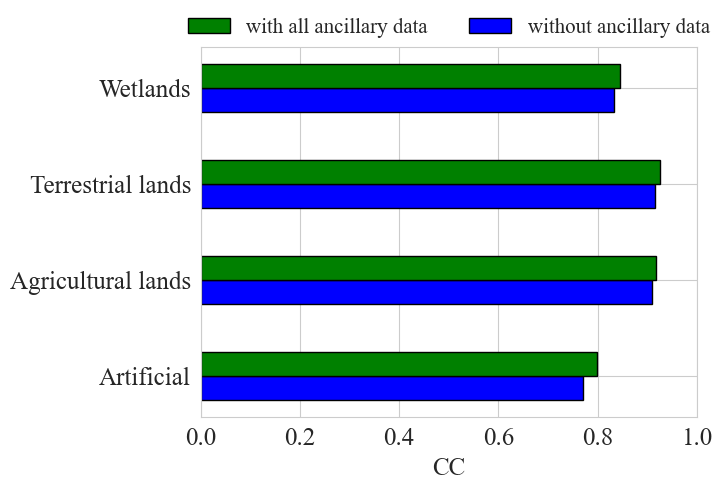}
    \label{fig:first}
\end{subfigure}
\begin{subfigure}{0.45\textwidth}
    \includegraphics[width=\textwidth]{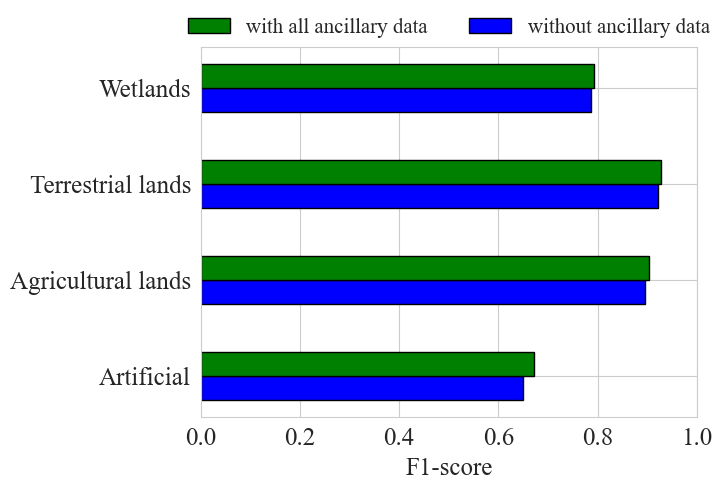}
    \label{fig:second}
\end{subfigure}
\begin{subfigure}{0.45\textwidth}
    \includegraphics[width=\textwidth]{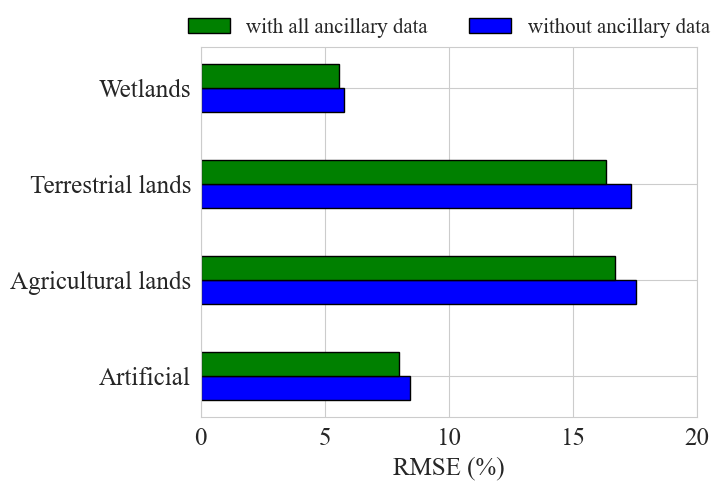}
    \label{fig:third}
\end{subfigure}
\begin{subfigure}{0.45\textwidth}
    \includegraphics[width=\textwidth]{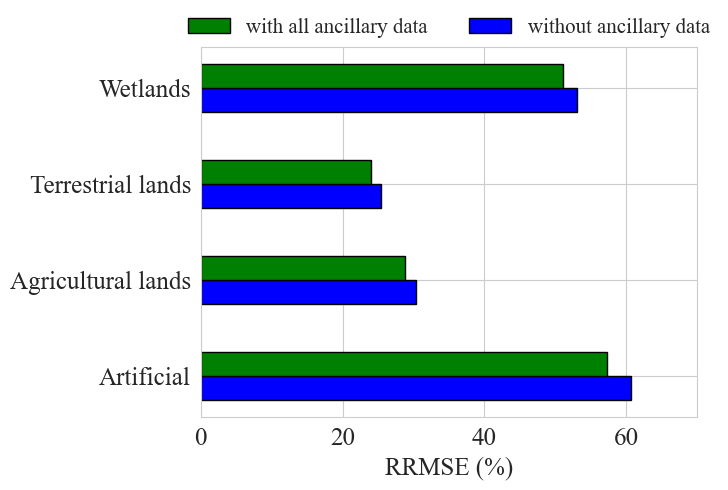}
    \label{fig:fith}
\end{subfigure}
\begin{subfigure}{0.45\textwidth}
    \includegraphics[width=\textwidth]{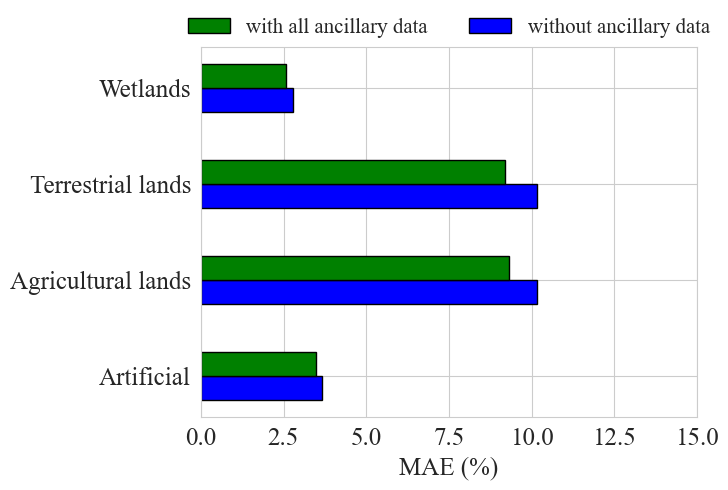}
    \label{fig:fith}
\end{subfigure}

\caption{Test results for the four SIPNA level 1 classes obtained by including all ancillary information (green) and without ancillary information (blue): CC values (top left), F1-score values (top right), RMSE values (middle left), RRMSE values (middle right), and MAE values (bottom).}
\label{fig:comparison_n1}
\end{figure*}

A further analysis of the five metrics for each class is depicted in Figure \ref{fig:comparison_n1}. In general, including the geo-topographic and climatic information improves the abundance predictions of all the classes of level 1. The "terrestrial lands" and "agricultural lands" achieve better performance in terms of CC, F1-score and RRMSE. However, the classes that benefit the most from adding the ancillary data are "artificial" and "wetlands" since the relative improvement is greater in these classes. 

It is worth noting that the RMSE and MAE metrics are not fair for comparisons between classes as they do not take into account the range of abundance values within each class. The most appropriate metric for these comparisons in this case is the RRMSE.

\begin{figure*}[]
\centering

\begin{subfigure}{0.49\textwidth}
\includegraphics[width=\textwidth]{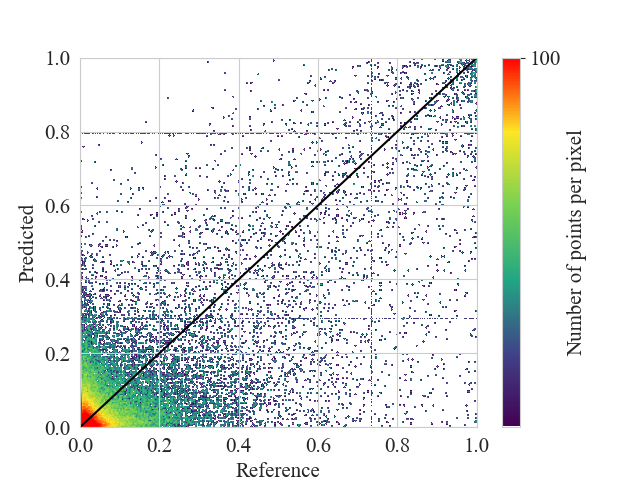}
\caption{Artificial}
\end{subfigure}
\begin{subfigure}{0.49\textwidth}
    \includegraphics[width=\textwidth]{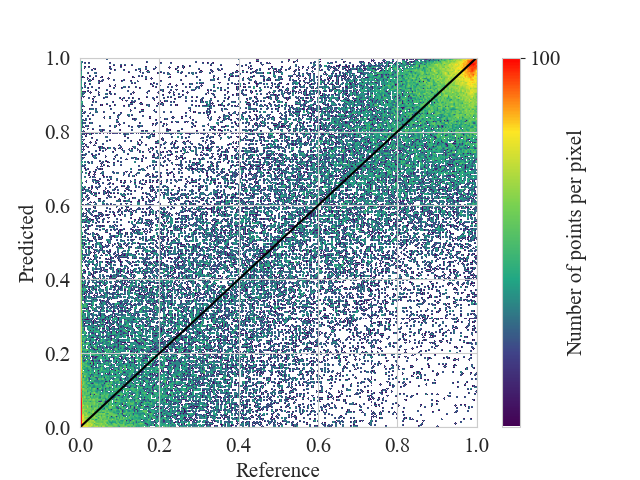}
    \caption{Agricultural lands}
\end{subfigure}
\begin{subfigure}{0.49\textwidth}
    \includegraphics[width=\textwidth]{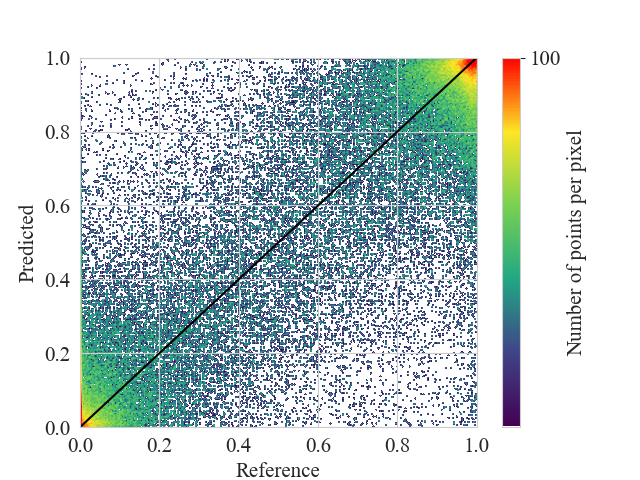}
    \caption{Terrestrial lands}
\end{subfigure}
\begin{subfigure}{0.49\textwidth}
    \includegraphics[width=\textwidth]{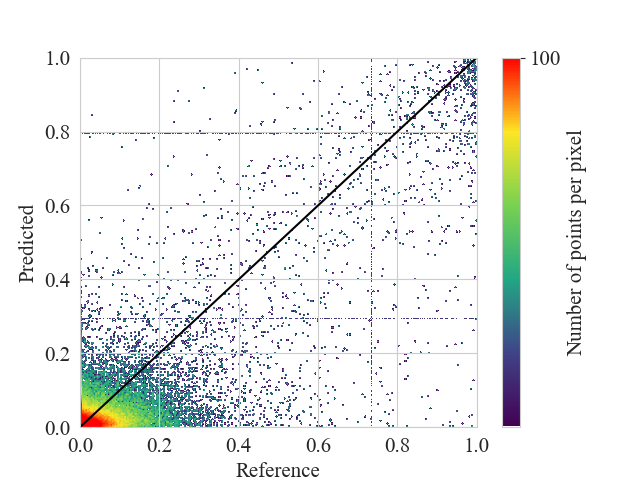}
    \caption{Wetlands}
\end{subfigure}
\vspace*{+2mm}
\caption{Density scatter plots of every level 1 class abundances (predicted vs reference) for the best model (including all ancillary data). (a) Artificial, (b) Agricultural lands, (c) Terrestrial lands, and (d) Wetlands.}
\label{fig:den_sca_n1}
\end{figure*}

To better illustrate the reasons behind these differences in performance between observed and predicted abundances in each of the four LULC classes, Figure \ref{fig:den_sca_n1} shows a density scatter plot for each class. The scatter plots of "artificial" and "wetlands" pixels showed a less aligned distribution along the 1:1 straight line than terrestrial and agricultural lands. In artificial and wetlands plots, most points are concentrated in the lowest abundances, while in terrestrial and agricultural lands points tend to concentrate in both the extremes of the abundance gradient but also along the 1:1 line. This proofs that the model works reasonably good for both abundant (terrestrial and agricultural lands) and scarce (artificial and wetlands) classes.

\begin{figure*}[h!]
    \centering
	\includegraphics[width=\textwidth]{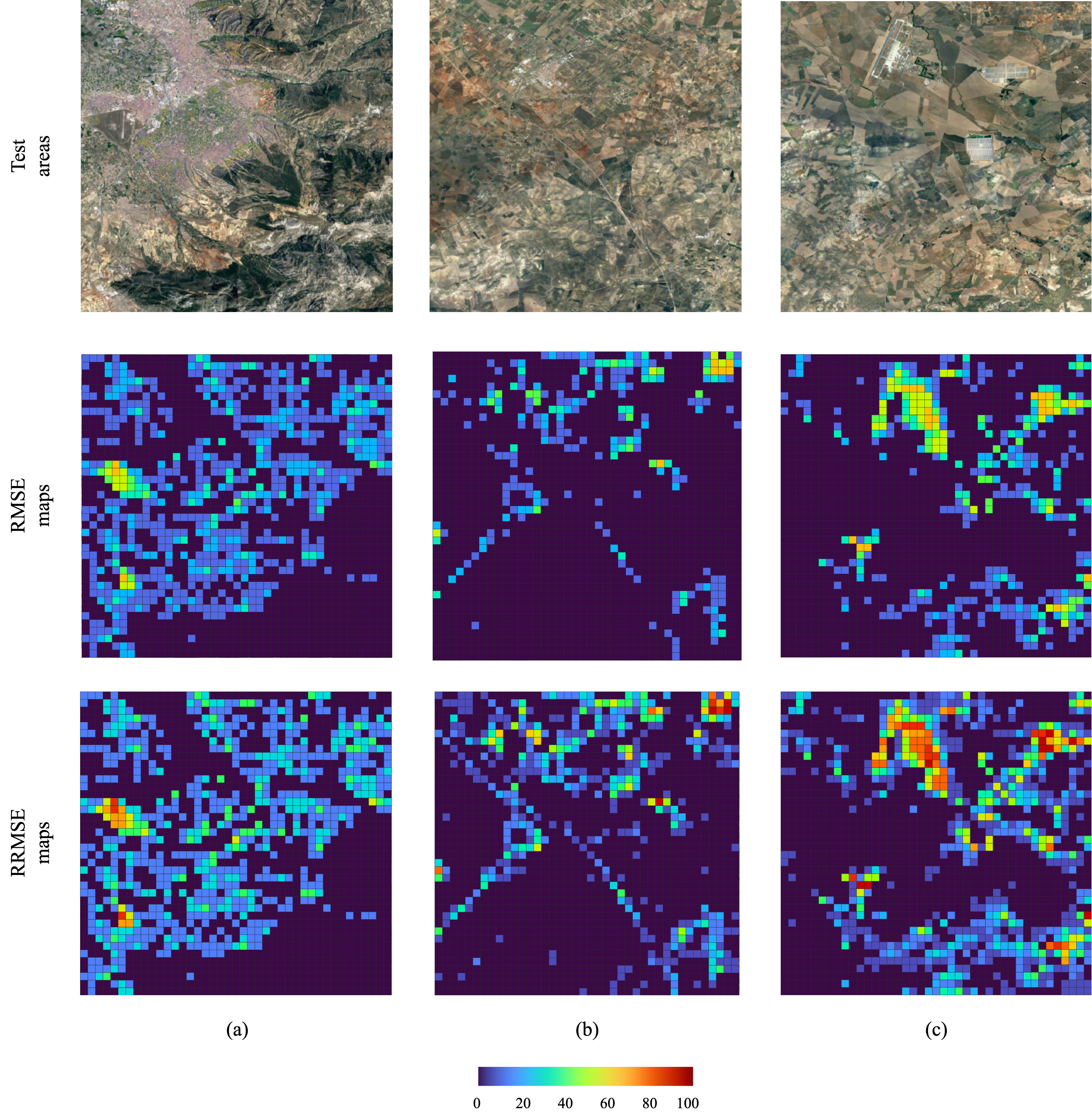}
	\caption{Three test areas (top row) with their corresponding RMSE (middle row) and RRMSE (bottom row) per pixel maps in level 1. (a) Granada, Granada, (b) La Carlota, Córdoba, (c) El Coronil, Sevilla.}
	\label{fig:n1_rmse_gradient}
\end{figure*}

Finally, Figure \ref{fig:n1_rmse_gradient} shows the results achieved by the best model on three test areas (top row) with their corresponding RMSE (middle row) and RRMSE (bottom row) per pixel maps. As we can observe, most of the pixels are in blue tones, which indicates a low RMSE and RRMSE and a great LULC abundances predictions. A reduced number of pixels with red tones in the RRMSE maps indicates an important prediction error relative to the scale of the reference abundance. These pixels mainly correspond to small heterogeneous rural areas with a large diversity of urban, crop and even forest areas, which makes the task of correctly predicting each and every LULC class abundances very difficult.

\subsection{SIPNA level 2}

Similarly, we evaluated the proposed model on these input combinations:  (spectro-temporal data), (spectro-temporal data + geo-topographic data),  (spectro-temporal data + climatic data) and (spectro-temporal data + geo-topographic and climatic data) considering SIPNA level 2. We also include the results of a baseline model trained from scratch on spectro-temporal data only to show how the pretraining on TimeSpec4LULC dataset is beneficial for level 2 as well. 
The results of these five models in terms of the average MAE, RMSE, RRMSE, CC, F1-score, RRMSE gain, CC gain across the ten classes of level 2 are provided in Table \ref{table:results_n2}. Aditionally, the computational complexity of each model is expressed in terms of MFLOPs in the last column. Again, we can see in the first two rows that by just pretraining our model in TimeSpec4LULC dataset improves the results in every metric, proving the value of pretraining DL models in similar tasks to achieve better performance. Similarly, including ancillary information improves the spectral unmixing task even in a much more difficult spectral unmixing setting (Table \ref{table:results_n2}). Compared to the baseline, the best performing model (including all the ancillary data) decreases the MAE, RMSE and RRMSE by 0.56\%, 0.65\% and 2.80\%, respectively and increases CC and  F1-score by 0.0320 and 0.0332, respectively.

\renewcommand{\arraystretch}{1.5}
\begin{table*}[h!]
\centering
\caption{Performance comparison of our model trained from scratch (first row) and finetuned from TimeSpec4LULC (second row) using only multi-spectral multi-temporal input data, by adding  geo-topographic data only (third row), by adding climatic data only (fourth row) and by adding both geo-topographic and climatic (fifth row). The performance is expressed in terms of average Mean Absolute Error (MAE), Root Mean Squared Error (RMSE), Relative Root Mean Squared Error (RRMSE), Correlation Coefficient (CC), F1-score, RMSE gain and CC gain with respect to baseline model for SIPNA level 2 classes. Last column, "MFLOPs", indicates the model's computational complexity in terms of Mega FLOPs.}
\vspace{+1mm}
\resizebox{0.98\textwidth}{!}{%
\begin{tabular}{l|r r r r r r r | r}
 \textbf{Ancillary data added} & \textbf{MAE (\%)} & \textbf{RMSE (\%)} & \textbf{RRMSE (\%)} & \textbf{CC} & \textbf{F1-score} & \textbf{RMSE (\%) gain} & \textbf{CC gain} & \textbf{MFLOPs}\\ \hline
 None (from scratch) & 6.13\% & 11.91\% & 60.23\% & 0.7372 & 0.6072  & 0.0000\% & 0.0000 & 2.3055 \\
 None (finetuned from TimeSpec4LULC)& 5.88\% & 11.54\% & 59.11\% & 0.7504 & 0.6216  & -0.3700\% & 0.0132 & 2.3055 \\
 Climatic & 5.73\% & 11.47\% & 58.57\% & 0.7566 & 0.6317  & -0.4400\% & 0.0233 & 2.3665 \\
 Geo-topographic & 5.60\% & 11.31\% & 57.92\% & 0.7635 & 0.6396  & -0.6000\% & 0.0312 & 2.3663 \\ 
 Geo-topographic + climatic & \textbf{5.57\%} & \textbf{11.26\%} & \textbf{57.43\%} & \textbf{0.7664} &  \textbf{0.6404}  &  \textbf{-0.6500\%} &  \textbf{0.0320} & 2.3673 \\ 
\end{tabular}
}
\label{table:results_n2}
\end{table*}
\renewcommand{\arraystretch}{1}

\begin{figure*}[h!]
\centering
\begin{subfigure}{0.49\textwidth}
    \includegraphics[width=\textwidth]{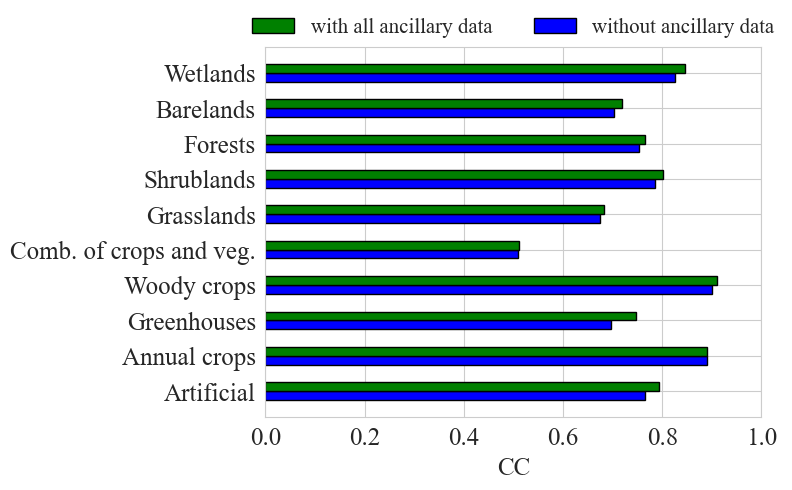}
\end{subfigure}
\hfill
\begin{subfigure}{0.49\textwidth}
    \includegraphics[width=\textwidth]{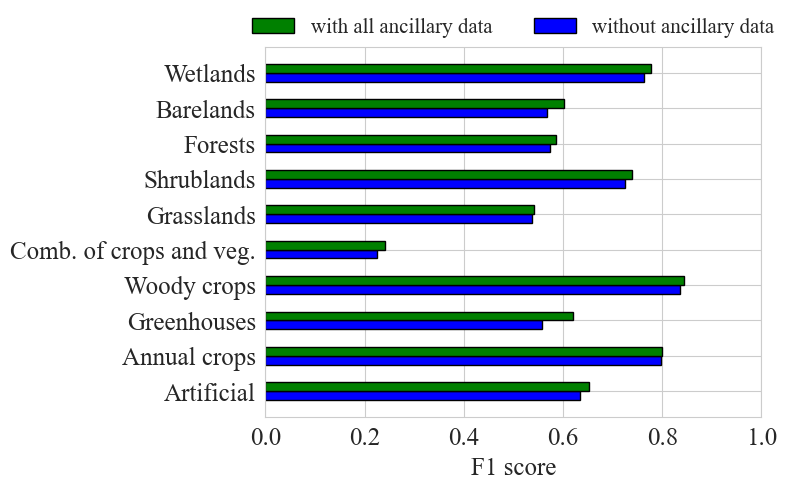}
\end{subfigure}
\hfill
\begin{subfigure}{0.49\textwidth}
    \includegraphics[width=\textwidth]{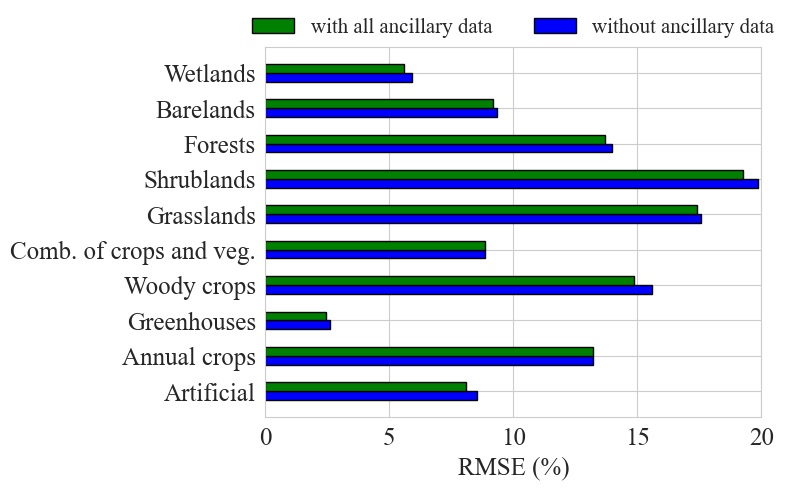}
\end{subfigure}
\hfill
\begin{subfigure}{0.49\textwidth}
    \includegraphics[width=\textwidth]{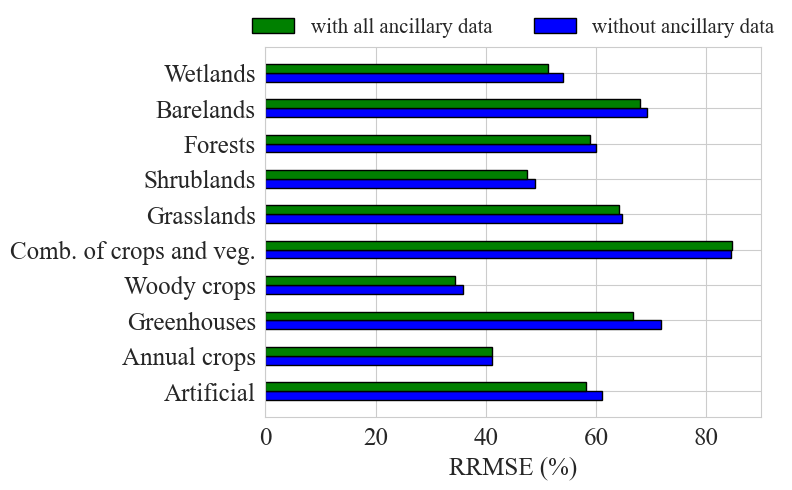}
\end{subfigure}
\hfill
\begin{subfigure}{0.49\textwidth}
    \includegraphics[width=\textwidth]{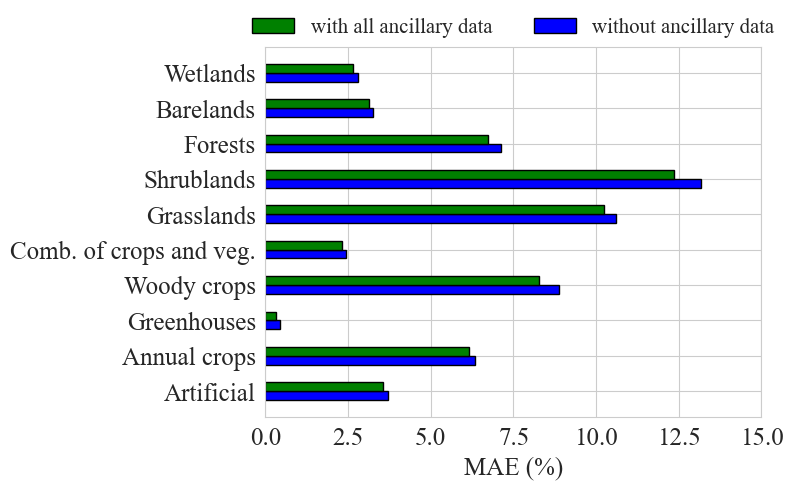}
\end{subfigure}

\caption{Test results for the ten SIPNA level 2 classes obtained by including all ancillary information (green) and without ancillary information (blue): CC values (top left), F1-score values (top right), RMSE values (middle left), RRMSE values (middle right), and MAE values (bottom).}
\label{fig:comparison_n2}
\end{figure*}

In the same way as in level 1, Figure \ref{fig:comparison_n2} shows a comparison between the baseline model and the model including geo-topographic and climatic data for every class in each of the five metrics used for evaluation. In general, adding ancillary information improves the abundances predictions of all the classes. The best performance is achieved in "woody crops" and "annual crops" classes in terms of CC, F1-score and RRMSE. Besides, adding the ancillary information to the model achieves a greater improvement for the classes with the worst results like "combinations of croplands and vegetation", "barelands" and "artificial". 

\begin{figure*}[h!]
\centering

\begin{subfigure}{0.32\textwidth}
\includegraphics[width=\textwidth]{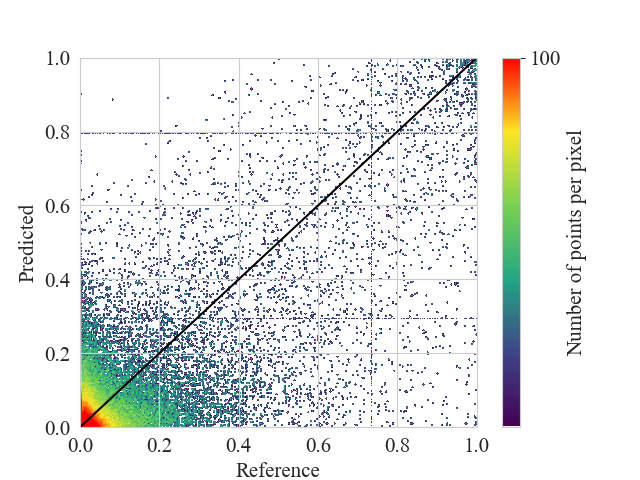}
\caption{Artificial}
\end{subfigure}
\hfill
\begin{subfigure}{0.32\textwidth}
    \includegraphics[width=\textwidth]{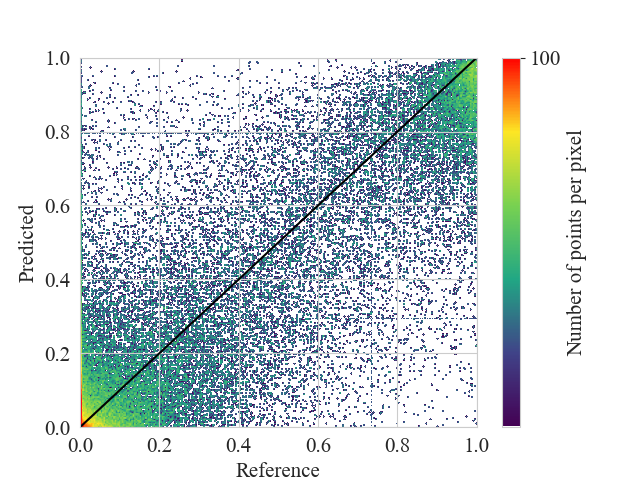}
    \caption{Annual crops}
\end{subfigure}
\hfill
\begin{subfigure}{0.32\textwidth}
    \includegraphics[width=\textwidth]{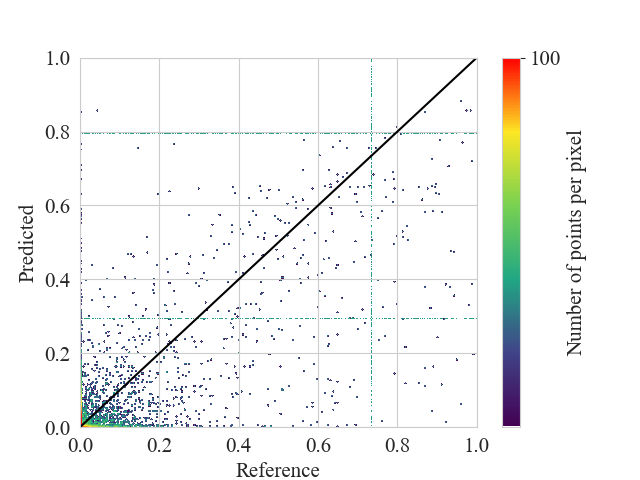}
    \caption{Greenhouses}
\end{subfigure}
\hfill
\begin{subfigure}{0.32\textwidth}
    \includegraphics[width=\textwidth]{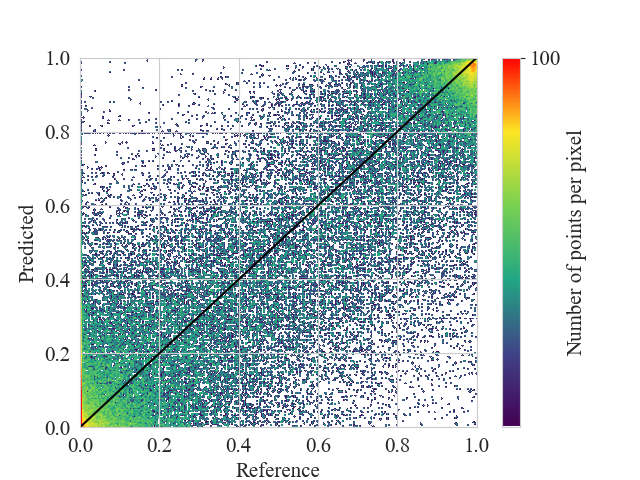}
    \caption{Woody crops}
\end{subfigure}
\hfill
\begin{subfigure}{0.32\textwidth}
    \includegraphics[width=\textwidth]{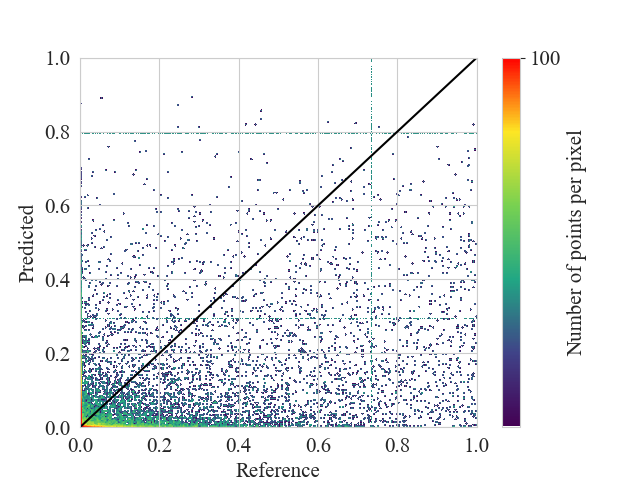}
    \caption{Crops \& vegetation}
\end{subfigure}
\hfill
\begin{subfigure}{0.32\textwidth}
    \includegraphics[width=\textwidth]{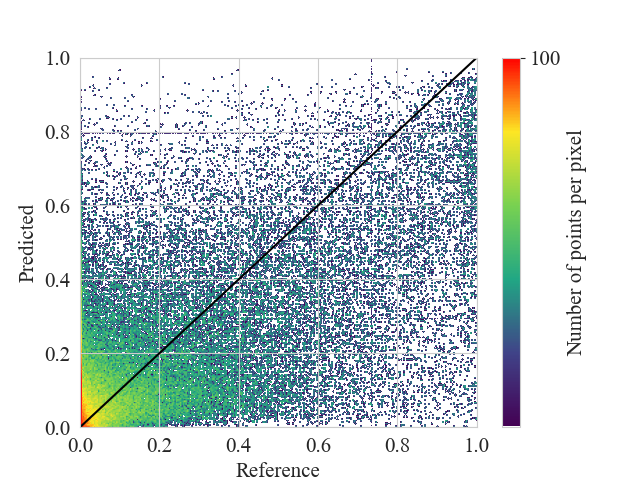}
    \caption{Grasslands}
\end{subfigure}
\hfill
\begin{subfigure}{0.32\textwidth}
    \includegraphics[width=\textwidth]{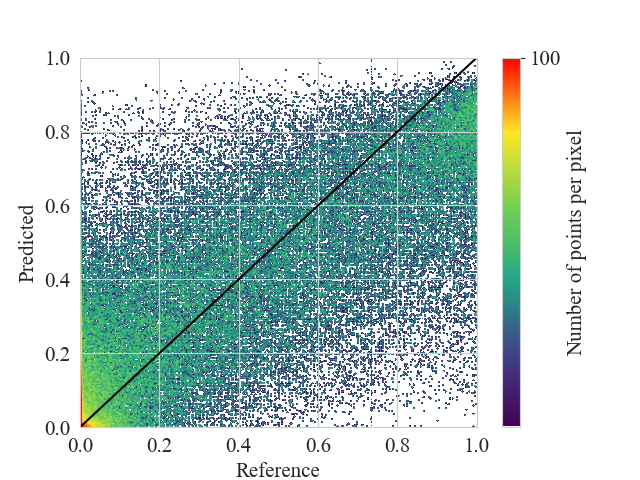}
    \caption{Shrublands}
\end{subfigure}
\hfill
\begin{subfigure}{0.32\textwidth}
    \includegraphics[width=\textwidth]{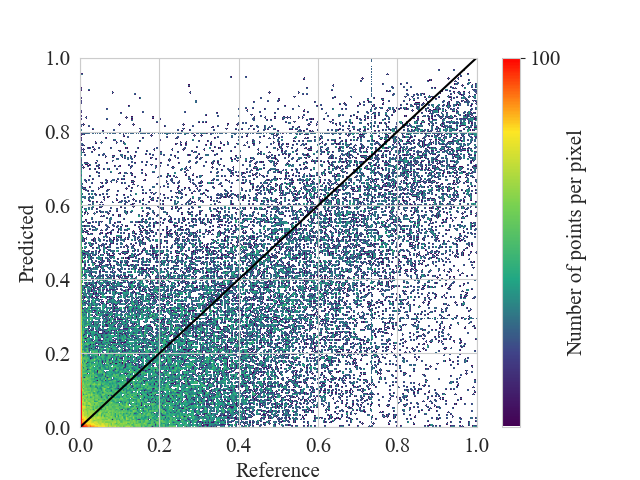}
    \caption{Forests}
\end{subfigure}
\hfill
\begin{subfigure}{0.32\textwidth}
    \includegraphics[width=\textwidth]{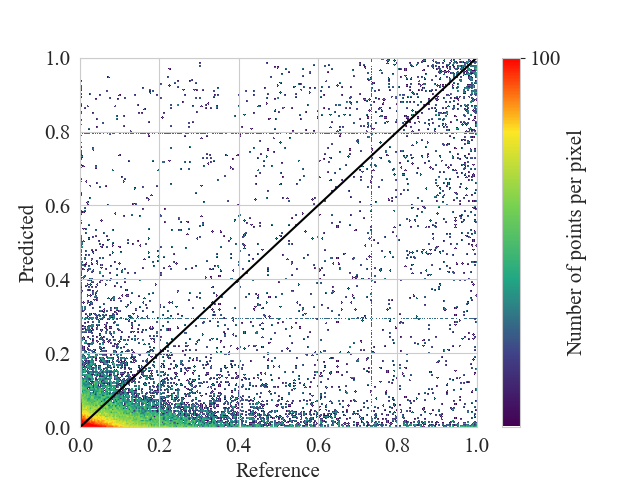}
    \caption{Barelands}
\end{subfigure}
\hfill
\begin{subfigure}{0.32\textwidth}
    \includegraphics[width=\textwidth]{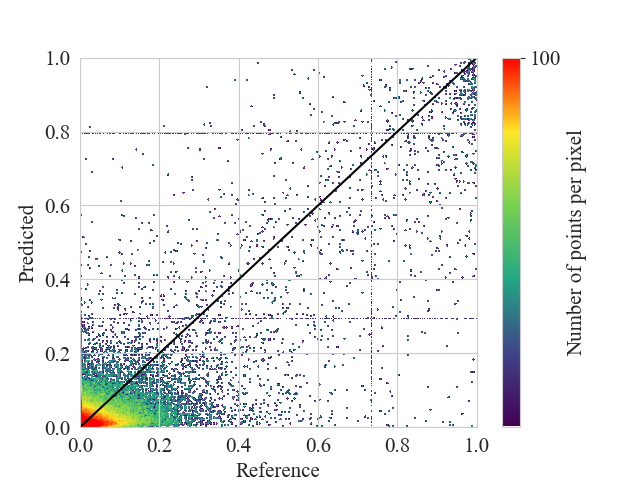}
    \caption{Wetlands}
\end{subfigure}
\vspace*{+2mm}
\caption{Density scatter plots of every level 2 class abundance (predicted vs reference) for the best model (including all ancillary data). (a) Artificial, (b) Annual croplands, (c) Greenhouses, (d) Woody croplands, (e) Combinations of croplands and vegetation , (f) Grasslands and grasslands with trees, (g) Shrublands and shrublands with trees, (h) Forests, (i) Barelands, and (j) Wetlands.}
\label{fig:den_sca_n2}
\end{figure*}

Looking at the density scatter plot for each level 2 class in Figure \ref{fig:den_sca_n2}, we see that the correlation between the reference and the predicted abundances is generally good, except for "combinations of croplands and vegetation" and "barelands" classes since they show a large dispersion. It is worth emphasizing the strong performance of the model for the class "greenhouses". Despite of having so few representation of middle range values of abundance in the pixels of Andalusia, the correlation in this class between the reference and predicted abundances is similar to the classes with a good representation. We argue that the reason for that could be due to their very high albedo, i.e., high reflectance in all bands. Finally, the worst performance metrics were obtained for  "combinations of croplands and vegetation" class, which may be due to the mixed-nature of this class definition itself. By combining crop and vegetation this class is a mixture of some of the other classes and hence it is complicated for the model to predict the correct abundances.

\begin{figure*}[h!]
    \centering
	\includegraphics[width=\textwidth]{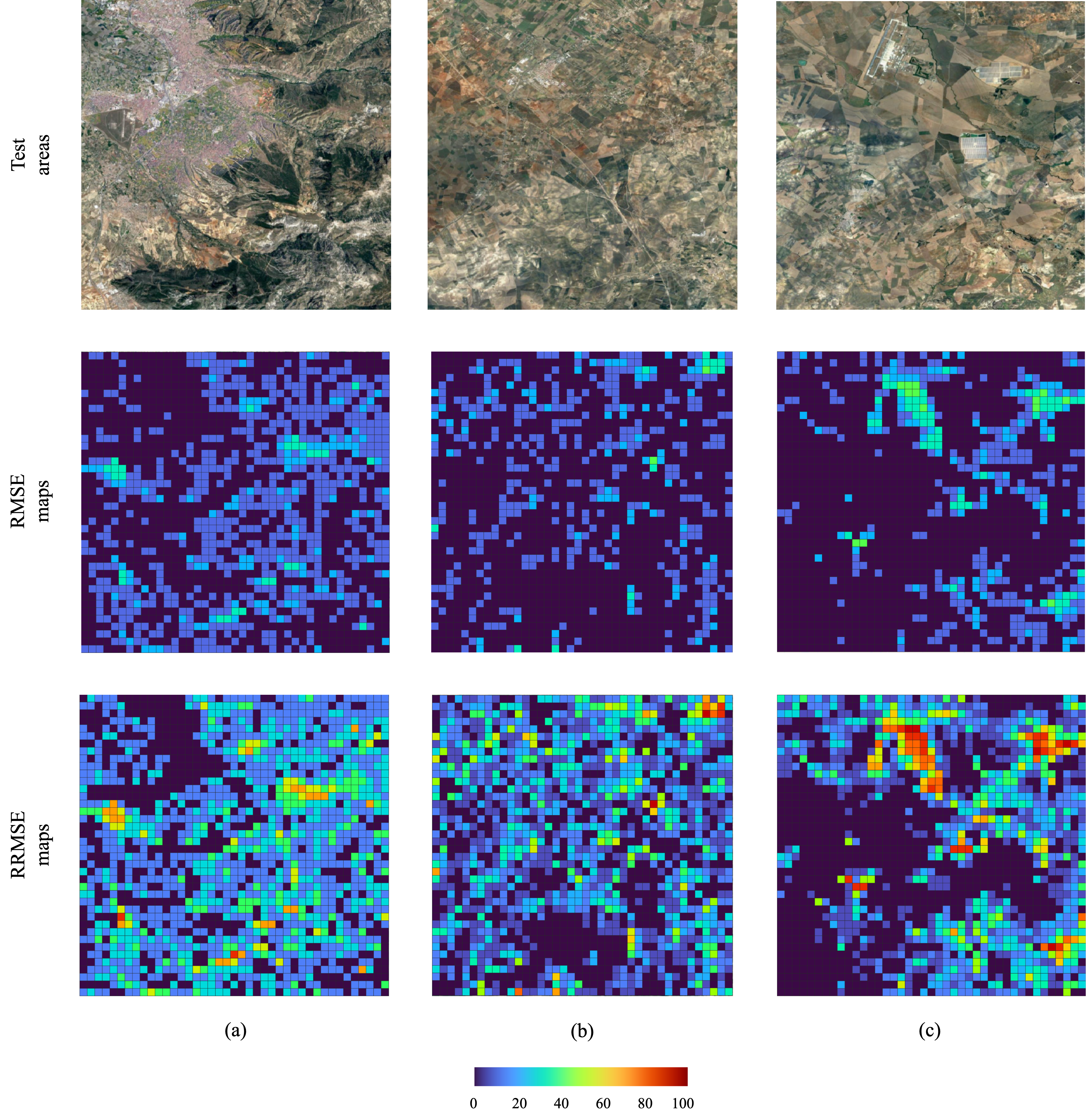}
	\caption{Three test areas (top row) with their corresponding RMSE (middle row) and RRMSE (bottom row) per pixel maps in level 2. (a) Granada, Granada, (b) La Carlota, Córdoba, (c) El Coronil, Sevilla.}
	\label{fig:n2_rmse_gradient}
\end{figure*}

Lastly, Figure \ref{fig:n2_rmse_gradient} shows the results achieved by the best model on three test areas (top row) with their corresponding RMSE (middle row) and  RRMSE (bottom row) per pixel maps. In general, most pixels are in dark blue  tones (low error) in the RMSE maps, which at first glance may seem better than the results achieved for level 1.  However, when looking at the RRMSE maps we can notice a slightly higher number of pixels with red tones  than in the level 1 RRMSE maps, located mainly in heterogeneous rural areas. Given that in level 2 we have 12 LULC classes, there are more heterogeneous pixels and consequently the unmixing task is harder. It is important to note that although the error at level 1 is lower in absolute terms, when it is relativized by the scale of the reference abundances it becomes higher than in level 1, indicating only moderate results compared to the good results obtained at level 1. For this reason, it is recommended to evaluate not only the RMSE but also the RRMSE to get better conclusions.

\section{Discussions} \label{sec:discussion}

Spectral unmixing of LULC classes is a challenging problem commonly addressed by physical models with the need for endmember extraction \cite{keshava2002spectral, 9324546}.  Moreover, the variability naturally present in the spectral signature for a given LULC class (spectral variability) makes this problem even more difficult \cite{8528557, 10015783}. DL methods represent a great solution to eliminate the need for endmember extraction and they are known for their robustness against noise given sufficient amounts of training data. Most previous works focus on HS or MS data and do not exploit  temporal information to estimate the abundance of mixed pixels. Obtaining MS time series of large territories can be prohibitive due to the cost and time required to acquire them  \cite{french, nalepa2021recent}. In addition, no previous work has explored the possibility of adding ancillary data to enhance the spectral unmixing results, which are used successfully in other computer vision tasks \cite{berg2014birdsnap, ellen2019improving, tseng2022timl}.

In this work, we tried to solve the mentioned constraints of previous works by: 

\begin{itemize}
    \item developing Andalusia-MSMTU, a high-quality MS time series dataset of mixed pixels labeled with LULC class abundances at two classification levels and making it publicly available so other researchers can develop new approaches in the field of spectral unmixing. This dataset followed several data pre-processing steps as explained in Section \ref{sec:data_extraction} in order to smooth spectral varibilities associated with the imaging process.

    \item proposing and analysing DL-based approaches for SU without the need of endmembers extraction. Moreover, we intentionally included pixels well distributed around our study area in models' training, which implies a high number of diverse pixels with different spectral variations. This way, the DL models will be robust against the spectral variations of pixels in the test areas.
\end{itemize}

Our results showed that our DL-based method achieved good results for spectral unmixing of LULC classes by using MS data and it can be used in areas with similar features such as the rest of Spain and mediterranean countries. Besides, by including ancillary information the model improved in terms of every metric used for evaluation, showing that adding external data is an interesting avenue to explore in spectral unmixing problems. 

Finally, one significant limitation still exists in our work. Although DL models have shown great performance in mapping complex input-output relationships and have demonstrated promising results for SU of LULC classes, they lack physical interpretability. This means that it is difficult to understand how the model arrived at its decision, and it may not be clear why certain input features were given more weight than others \cite{arrieta2020explainable}. In the context of spectral unmixing, physical interpretation may be desirable \cite{9554521} because it allows us to understand the underlying physical processes that govern the interaction of electromagnetic radiation with land surface materials.

\section{Conclusions} \label{sec:conclusions}

In this work, we introduced and made publicly available Andalusia-MSMTU dataset, a new DL-ready dataset to explore SU approaches on MS time series data. Furthermore, we introduced ancillary information to improve the spectral unmixing performance consisting on two geographic, two topographic and five climatic variables. Our experiments show that the use of MS time series data for LULC abundance estimation achieves good results, which are further improved by including ancillary information.

For future work, we would like to explore taking advantage of spatial autocorrelation between neighbouring pixels, which provides useful information for the spectral unmixing task \cite{shi2014incorporating}, by arranging the MODIS pixels in images and using a Convolutional-LTSM network with a BRITS-like approach to deal with missing values. Moreover, given the recent availability of higher spatial resolution sensors like Sentinel-2, data fusion between MODIS long-term data and Sentinel-2 higher resolution data is another avenue to improve spectral unmixing performance. Finally, since common DL-based models lack physical interpretation and it is sometimes important in the context of spectral unmixing, an effort to make DL-based methods physically aware is worthwhile.

\section*{Acknowledgment}
This research was developed as part of the project EarthCul PID2020-118041GB-I00 within the Spanish Research Projects Plan funded by MCIN/AEI/10.13039/501100011033 and by FEDER funds “Una manera de hacer Europa”. This study also occurred thanks to the projects ECOPOTENTIAL, which received funding from the European Union’s Horizon 2020 Research and Innovation Programme under agreement No. 641762, and by the TED project with reference TED2021-129690B-I00 funded by the Ministry of Science and Innovation. Finally, this work was partially supported by the project “Thematic Center on Mountain Ecosystem \& Remote sensing, Deep learning-AI e-Services University of Granada-Sierra Nevada” (LifeWatch-2019-10-UGR-01), which has been co-funded by the Ministry of Science and Innovation through the FEDER funds from the Spanish Pluriregional Operational Program 2014-2020 (POPE), LifeWatch-ERIC action line. Additionally, we would like to express our sincere gratitude to Yassir Benhammou for his expertise into utilizing GEE, and Carlos Navarro and Thedmer Postma for their support in the data preprocessing in the reference LULC abundances computation, which contributed to the success of this study.
\bibliographystyle{unsrt}  
\bibliography{references}

\end{document}